\documentclass[10pt,twocolumn,letterpaper]{article}

\usepackage[pagenumbers]{cvpr} 

\usepackage{graphicx}
\usepackage{amsmath}
\usepackage{amssymb}
\usepackage{booktabs}
\usepackage{subcaption}
\usepackage{algorithm}
\usepackage{algorithmicx}
\usepackage{algpseudocode}
\usepackage{color}
\usepackage{multirow}
\usepackage[fleqn]{nccmath}

\usepackage[pagebackref,breaklinks,colorlinks]{hyperref}

\newcommand{\xia}{\check{\mathbf{x}}}
\newcommand{\xa}{\hat{\mathbf{x}}}
\newcommand{\x}{\mathbf{x}}
\newcommand{\z}{\mathbf{z}}

\algnewcommand\algorithmicforeach{\textbf{for each}}
\algdef{S}[FOR]{ForEach}[1]{\algorithmicforeach\ #1\ \algorithmicdo}

\usepackage[capitalize]{cleveref}
\crefname{section}{Sec.}{Secs.}
\Crefname{section}{Section}{Sections}
\Crefname{table}{Table}{Tables}
\crefname{table}{Tab.}{Tabs.}

\def\cvprPaperID{*****} 
\def\confName{CVPR}
\def\confYear{2023}

\begin{document}

\title{The Enemy of My Enemy is My Friend: \\ Exploring Inverse Adversaries for Improving Adversarial Training}

\author{Junhao Dong\textsuperscript{1}, Seyed-Mohsen Moosavi-Dezfooli\textsuperscript{2}, Jianhuang Lai\textsuperscript{1} and Xiaohua Xie\textsuperscript{1}\\
	\textsuperscript{1}Sun Yat-Sen University, China,~~
	\textsuperscript{2}Imperial College London, UK\\
	{\tt\small dongjh8@mail2.sysu.edu.cn, seyed.moosavi@imperial.ac.uk,} \\ {\tt\small \{stsljh, xiexiaoh6\}@mail.sysu.edu.cn}
}

\maketitle

\begin{abstract}
Although current deep learning techniques have yielded superior performance on various computer vision tasks, yet they are still vulnerable to adversarial examples. Adversarial training and its variants have been shown to be the most effective approaches to defend against adversarial examples. These methods usually regularize the difference between output probabilities for an adversarial and its corresponding natural example. However, it may have a negative impact if the model misclassifies a natural example. To circumvent this issue, we propose a novel adversarial training scheme that encourages the model to produce similar outputs for an adversarial example and its ``inverse adversarial'' counterpart. These samples are generated to maximize the likelihood in the neighborhood of natural examples. Extensive experiments on various vision datasets and architectures demonstrate that our training method achieves state-of-the-art robustness as well as natural accuracy. Furthermore, using a universal version of inverse adversarial examples, we improve the performance of single-step adversarial training techniques at a low computational cost.
\end{abstract}

\vspace{-3mm}
\section{Introduction}

\label{sec:intro}

Deep learning has achieved revolutionary progress in numerous computer vision tasks \cite{SimonyanZ14a, long2015fully, zhao2019object} and has emerged as a promising technique for fundamental research in multiple disciplines \cite{ronneberger2015u, piloto2022intuitive, zemskova2022deep}. However, a well-established study has demonstrated that Deep Neural Networks (DNNs) are extremely vulnerable to adversarial examples \cite{SzegedyZSBEGF13, GoodfellowSS14}, which are indistinguishable from natural examples in human vision. In other words, a visually undetectable perturbation to the original example can lead to a significant disruption of the inference result of DNNs. The stealthiness of these tailored examples also makes them easy to bypass manual verification \cite{carlini2017adversarial, hendrycks2021natural}, posing a potential security threat to the safety of deep learning-based applications. Consequently, adversarial robustness has been considered a new measurement for deep learning security.

\begin{figure}[t]
	\centering
	\begin{subfigure}[t]{0.49\linewidth} 
		\centering
		\includegraphics[width=1\linewidth]{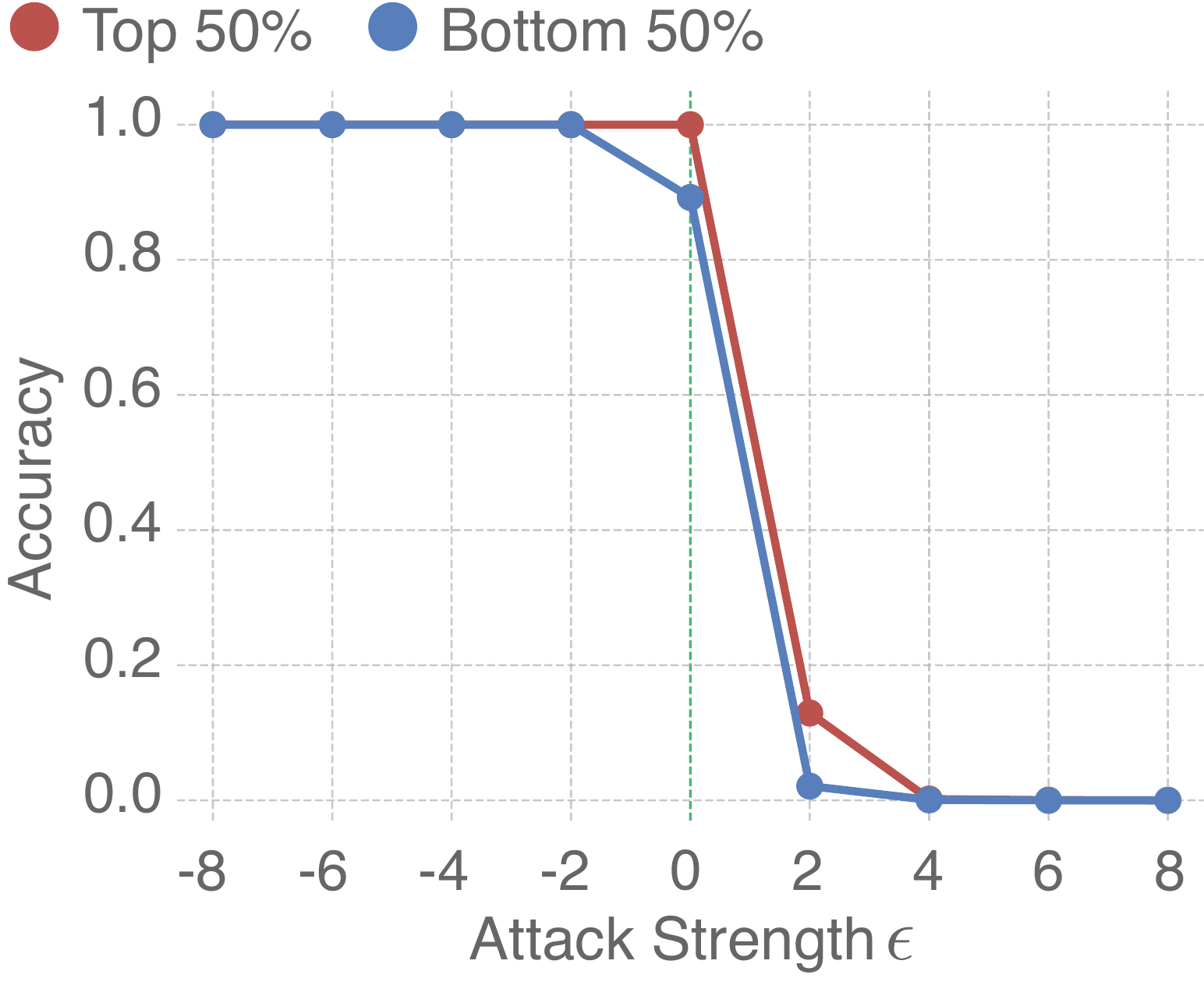}
		\caption{Natural Training}
		\label{fig:1_1}
	\end{subfigure} 
	\begin{subfigure}[t]{0.49\linewidth}
		\centering
		\includegraphics[width=1\linewidth]{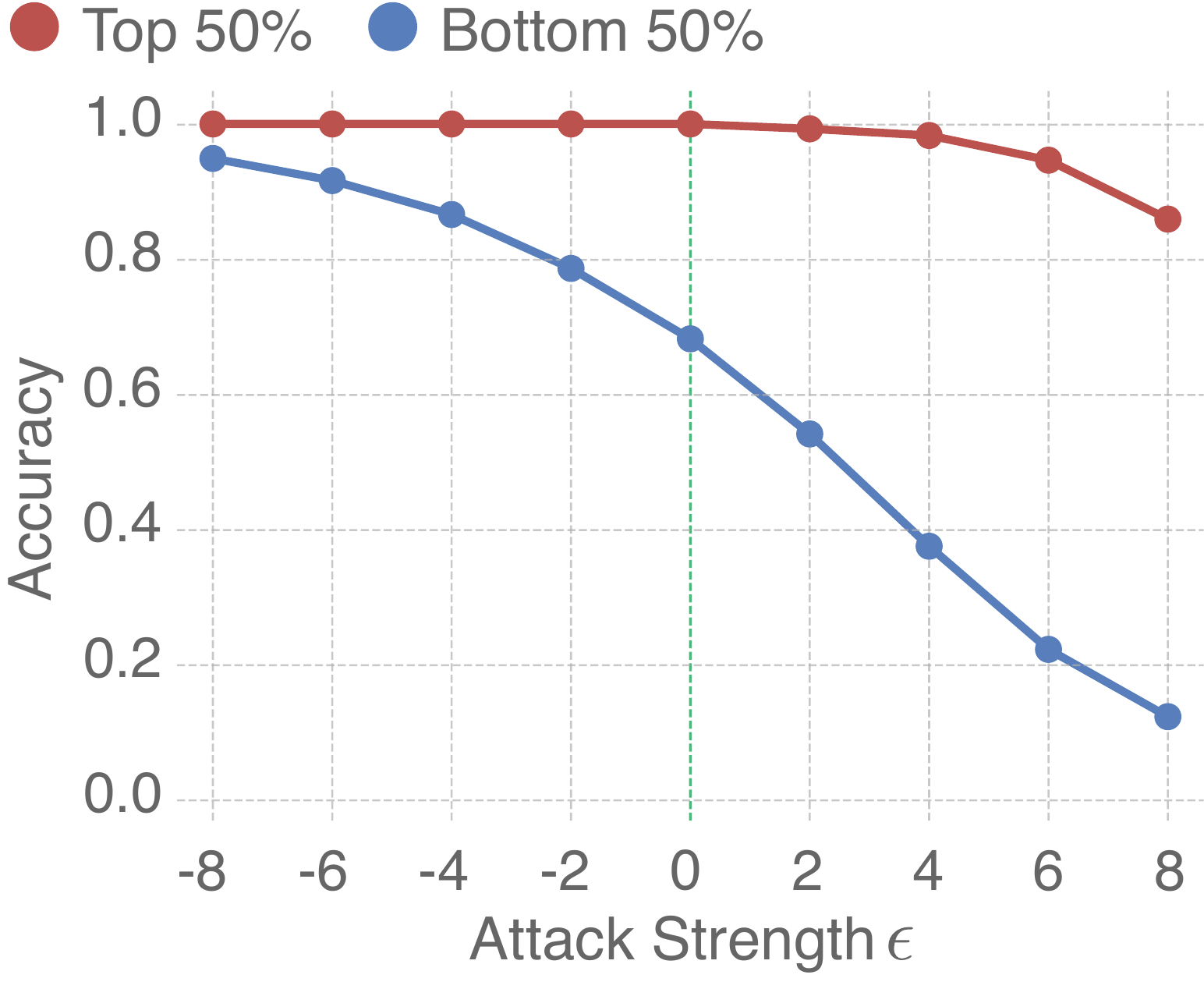}
		\caption{Adversarial Training \cite{MadryMSTV18}}
		\label{fig:1_2}
	\end{subfigure} 
	\vspace{-1mm}
	\caption{Average accuracy under different attack strengths for two networks trained on natural and adversarial samples. We rank test examples based on the natural classification accuracy in decreasing order and divide them into two equal halves. Note that the negative $\epsilon$ denotes the strength of inverse adversarial perturbation. (a) Naturally trained models are extremely susceptible to perturbations. (b) For adversarially trained models, the adversarial effect is exacerbated on examples that are more possibly to be misclassified. The \textcolor[RGB]{113,177,134}{green} line corresponds to natural examples.
	}
	\label{fig:1}
	\vspace{-1mm}
\end{figure}

Various defense methods have been proposed to improve the adversarial robustness of DNNs \cite{XieWZRY18, wong2018provable, lee2018simple}. As the primary defense method, adversarial training \cite{SzegedyZSBEGF13, GoodfellowSS14, MadryMSTV18} improves intrinsic network robustness via adaptively augmenting adversarial examples into training examples. Existing adversarial training methods mainly focus on the distribution alignment between legitimate examples and adversarial examples to preserve the consistency of the DNN prediction results \cite{zhang2019theoretically, wang2019improving, cui2021learnable}. However, there still exists a considerable feature representation gap between natural examples and their adversarial counterparts, resulting in the undesirable decision boundary for the misclassification of natural examples. The misclassification can further be exacerbated when confronted with adversarial examples.

The natural intuition is that: most adversarial examples corresponding to misclassified natural examples are more possibly to be misclassified. Opposite to adversaries that are harmful to DNNs, we introduce inverse adversarial examples\footnote{The formal definition will be given in the following sections.} that are created via minimizing the objective function as an inverse procedure of adversary generation. Specifically, inverse adversarial examples are affiliative to DNNs, which can be more possibly to be correctly classified. To support this claim, we study the accuracy of trained classification models on two groups of samples (see Figure \ref{fig:1}). We present the accuracy of adversarial examples and their inverse counterparts under different attack strengths. For the adversarially trained model, the robust accuracy of examples with high loss suffers from a heavier drop than that of examples with low loss under larger attack strengths. This means that the adversarial counterparts of low-confidence or even misclassified examples are also misclassified. Therefore, the distribution alignment between two misclassified examples might have an unnecessary or even harmful effect on the robustness establishment.

In this paper, beyond the unnecessary or even harmful matching manner between misclassified examples, we propose a novel adversarial training framework based on the inverse version of adversarial examples, dubbed \textit{Inverse Adversarial Training} (IAT), which implicitly bridges the gap between adversarial examples and the high-likelihood region of their belonging classes. Adversarial examples of a certain category can thus be pulled closer to the high-likelihood region instead of their original examples. Specifically, we involve an inverse procedure of the standard adversary generation to obtain the high-likelihood region. In general, inverse adversarial examples can be viewed as the regularization of original examples for reducing prediction errors. Considering the multi-class decision surface and computational cost, we design a class-specific inverse adversary generation paradigm as opposed to the instance-wise pattern. Furthermore, we establish a momentum mechanism for the prediction of inverse adversaries to stabilize the training process. A one-off version of the inverse adversary generation is also proposed for improving time efficiency.

Comprehensive experiments demonstrate the effectiveness and generalizability of our method that can efficiently obtain a better trade-off between natural accuracy and robustness. We also show that our method can also be adapted to larger models with extra generated data for robustness enhancement. Besides, the robustness of single-step adversarial training methods can be further improved at a low cost by incorporating our method.

The main contribution of this paper can be summarized as follows:
\begin{itemize}
	
	\item By analyzing the unnecessary, or even harmful, alignment between misclassified examples, we propose a novel adversarial training framework based on the inverse version of adversarial examples, which promotes the aggregation of adversarial examples to the high-likelihood region of their belonging classes.
	\item Based on the proposed \textit{Inverse Adversarial Training} (IAT) paradigm, we further design a class-specific universal inverse adversary generation strategy to mitigate the class-wise imbalance hiding in the decision surface. We also propose a one-off inverse adversary generation strategy to reduce computational costs with a negligible performance loss.
	\item Extensive experiments demonstrate the effectiveness of IAT compared with state-of-the-art methods when using large models with extra synthetic data. Furthermore, we achieve a better trade-off between natural accuracy and adversarial robustness efficiently. Our method can also be combined with single-step adversarial training methods as a plug-and-play component for boosting robustness at a low cost.
\end{itemize}

\paragraph{Related Works.} The lethal vulnerabilities of deep neural networks against adversarial examples have been witnessed in \cite{SzegedyZSBEGF13, GoodfellowSS14, moosavi2016deepfool, carlini2017towards}. A myriad of attempts have been made to defend against these tailored examples, including adversarial training \cite{MadryMSTV18, zhang2019theoretically, wang2019improving, jia2022adversarial}, adversarial detection \cite{HendrycksG17a, tian2018detecting}, and input transformation-based methods \cite{XieWZRY18, SamangoueiKC18, yoon2021adversarial}. Among them, adversarial training consistently remains to be the most effective method \cite{BaiL0WW21} to improve intrinsic network robustness via augmenting the training data with adversarial examples. In addition, most existing works generally incorporate a regularization term to narrow the distribution difference between natural examples and their adversarial counterparts \cite{zhang2019theoretically, wang2019improving, cui2021learnable}, which has been demonstrated to be beneficial for robustness enhancement. This matching manner seems natural but might be misguided by misclassified natural examples, as we showed in Figure \ref{fig:1}. To circumvent this issue, several efforts have been devoted to assigning weights on losses in terms of the intensity of adversarial examples \cite{ZhangZ00SK21, liu2021probabilistic, dong2022improving}. However, they mainly concentrate on mitigating the imbalance of disturbance effect among adversarial examples, while our primary focus is to prevent the harmful alignment between misclassified examples by incorporating inverse adversarial examples.

Inverse adversarial examples were first formally described in \cite{salman2021unadversarial}, where Salman \textit{et al.} studied them in vision systems to enhance in-distribution performance against new corruptions. In comparison, we investigate the regularization effect of inverse adversarial examples on the distribution alignment during adversarial training for robustness enhancement. A concurrent work \cite{li2022collaborative} also exploits the inverse version of adversarial examples for adversarial robustness incorporating different distance metrics. However, we built on class-specific universal inverse adversaries for adversarial training with more efficiency and robustness. We also involve the feature-level prior knowledge in the inverse adversary generation for supplementary regularization. Furthermore, we show how our method can be combined with single-step adversarial training techniques to improve both the natural performance and robustness.

\section{Background}
\label{sec:background}
Consider a DNN classifier $f_{\boldsymbol{\theta}}: \mathcal{X} \rightarrow \mathbb{R}^{C}$ with parameters $\boldsymbol{\theta}$ that predicts probabilities of $C$ classes. Specially, we symbolize the output feature representation of the penultimate layer (before logits) $\mathcal{F}_{\boldsymbol{\theta}}(\mathbf{x})$ for a given example $\mathbf{x} \in \mathcal{X}$. Adversarial training can be an effective way to enhance the robustness of DNNs against adversarial perturbations, which adaptively involves adversarial examples in training as strong data augmentation. For a specific dataset $\left( \mathbf{x}, y\right)\sim \mathcal{D}$, the standard adversarial training \cite{MadryMSTV18} against attacks under $\ell_{\infty}$-norm threat model can be formulated as the following min-max optimization problem:

\begin{equation}
	\begin{aligned}
		\min\limits_{\boldsymbol{\theta}} \mathbb{E}_{\left( \mathbf{x}, y\right)\sim \mathcal{D} }\left[ \max\limits_{\left\| \boldsymbol{\delta} \right\|_{\infty}<\epsilon}\mathcal{L}_{\text{CE}}\left( f_{\boldsymbol{\theta}} \left( \mathbf{x}+\boldsymbol{\delta}\right) , y \right)  \right],
		\label{eq:1}
	\end{aligned}
\end{equation}
where $\mathcal{L}_{\text{CE}}$ is the cross-entropy loss and $\delta$ is the adversarial perturbation under the $\ell_{\infty}$-norm bound $\epsilon$. The outer minimization is to optimize empirical adversarial risk over the network parameters $\boldsymbol{\theta}$. The inner maximization of adversarial training can be viewed as searching for the most harmful adversarial examples $\mathbf{\hat{x}} = \mathbf{x} + \delta$, which can be simplified as an iterative Projected Gradient Descent (PGD) algorithm \cite{MadryMSTV18} on the negative loss function.

Besides standard adversarial training, TRADES\cite{zhang2019theoretically} and MART\cite{wang2019improving} proposed to utilize Kullback–Leibler (KL) divergence for distribution matching between natural examples and their adversarial counterparts. The objective function of TRADES \cite{zhang2019theoretically} can be defined as follows:

\begin{equation}
	\begin{aligned}
		\min\limits_{\boldsymbol{\theta}} \mathbb{E}_{\left( \mathbf{x}, y\right)\sim \mathcal{D} }  \Big[ & \mathcal{L}_{\text{CE}} \left( f_{\boldsymbol{\theta}} \left( \mathbf{x}\right) , y \right) + \\
		 \omega & \cdot  \max\limits_{\left\| \delta \right\|_{\infty}<\epsilon}\mathcal{L}_{\text{KL}}\left( f_{\boldsymbol{\theta}} \left( \mathbf{x}\right)  \| f_{\boldsymbol{\theta}}\left( \mathbf{x} + \delta \right) \right)   \Big],
		\label{eq:2}
	\end{aligned}
\end{equation}
where $\mathcal{L}_{\text{KL}}$ denotes KL divergence and $\omega$ is the balancing parameter for the trade-off of natural accuracy and adversarial robustness. Generally, KL divergence encourages the predictions of benign examples and adversarial examples to share the same distribution. Nevertheless, this distribution alignment can further undermine the adversarial robustness when benign examples are misclassified, resulting in the wrong guidance during adversarial training. A major drawback in adversarial training is that it costs more considerable computing resources than natural training \cite{shafahi2019adversarial, WongRK20, andriushchenko2020understanding}, which hinders robust establishment on larger models. In addition, adversarial training can suffer from more severe overfitting than the natural training paradigm \cite{rice2020overfitting}. Later in this paper, we will provide some insights related to the above-mentioned challenges regarding adversarial training.

\section{Method}
\label{sec:method}
In this section, we first formally define the inverse adversarial example and introduce its class-specific (universal) variant. We then propose a new adversarial training scheme, coined as \textit{Universal Inverse Adversarial Training} (UIAT), via a regularizer that encourages the prediction alignment between adversarial examples and the high-likelihood region of their corresponding classes. Furthermore, the inverse adversary momentum is also proposed for the stabilization of the training process. For boosting time efficiency, we design a one-off version of UIAT by computing inverse adversaries only in one of the epochs without losing much performance.

\subsection{Inverse Adversarial Examples}
For the image classification task, the adversary generation can be viewed as a process of crossing the decision boundary for misclassification. On the contrary, generating inverse adversarial examples can be regarded as moving away from the decision boundary to the high-likelihood region of a certain class. Specifically, this process can be obtained by iteratively minimizing the classification loss values of inverse adversarial examples. Formally, inverse adversarial examples are inputs to machine learning models, which are tailored to cause the model to obtain more accurate predictions than corresponding natural examples. Similar to adversarial examples, inverse adversaries are obtained by adding visually tiny perturbations to natural ones. We here focus on $\ell_{\infty}$-norm bound $\mathbb{B}(\mathbf{x}, \epsilon')$ with radius $\epsilon'$ around natural examples on inverse adversaries. One can use PGD to generate inverse adversarial perturbations: 

\begin{equation}
	\begin{aligned}
		\xia^{t+1} \!=\! \Pi_{\mathbb{B}(\mathbf{x}, \epsilon')} \!\left(  \xia^{t} \!-\! \alpha' \cdot \operatorname{sign} \left( \nabla_{\xia^{t}}\mathcal{L}_{\text{Inv}}\left( \xia^{t}, y \right)  \right)  \right)\!,
		\label{eq:3}
	\end{aligned}
\end{equation}
where $\alpha'$ is the gradient descent step size, $\xia^t$ represents $t^\text{th}$ iteration update, and $\mathcal{L}_{\text{Inv}}$ denotes the loss function for the inverse adversary generation. Generally, the cross-entropy loss can be a good choice for guiding the inverse adversary generation. Nevertheless, the high-likelihood region of a certain class is far away from any adjacent decision boundaries \cite{wen2016discriminative, kahla2022label}, which means that inverse adversaries are far away from adversarial examples at the feature level. Meanwhile, natural feature embeddings are also desired to lie on the high-likelihood region from the geometric perspective. We thus append a feature-level regularization during the inverse adversary generation for supplementary supervision. Therefore, given a sample $\x$ and its adversarial counterpart $\xa$, our inverse adversarial loss can be written as follows:

\begin{equation}
	\begin{aligned}
		\mathcal{L}_{\text{Inv}}\left( \check{\mathbf{x}}, y \right)=&\mathcal{L}_{\text{CE}}\left(f_{\boldsymbol{\theta}}\left(\xia \right), y \right) \\
		+ \beta \cdot [  \mathcal{L}_{\text{1}}\left( \mathcal{F}_{\boldsymbol{\theta}}\left(\xia \right), \mathcal{F}_{\boldsymbol{\theta}}\left(\mathbf{x} \right) \right) &- \mathcal{L}_{\text{1}}\left( \mathcal{F}_{\boldsymbol{\theta}}\left(\xia \right), \mathcal{F}_{\boldsymbol{\theta}}\left(\mathbf{\hat{x}} \right) \right) ],
		\label{eq:4}
	\end{aligned}
\end{equation}
where $\beta$ denotes the weighting factor. Similar to insights from adversarial feature space analysis\cite{mao2019metric}, our triplet term on latent representations can further prevent the overfitting of inverse adversaries from extremely high predictions for better guidance. The obtained inverse adversarial examples can then be incorporated into the adversarial training for robustness enhancement.

\begin{algorithm*}[tb] 
	\caption{\textbf{U}niversal \textbf{I}nverse \textbf{A}dversarial \textbf{T}raining (\textbf{UIAT})}  
	\label{alg:1} 
	\begin{algorithmic}[1] 
		\Statex {\bfseries Input:}  
		DNN classifier $f_{\boldsymbol{\theta}}$; dataset $\mathcal{D} = \{(\mathbf{x}, y)\}$ with $C$ classes; batch size m; learning rate $\tau$; radius for adversaries $\epsilon$ and inverse adversaries $\epsilon'$; step size $\alpha'$ for inverse adversary generation; weighting factors $\lambda$; momentum factor $\gamma$. 
		\State Randomly initialize the network parameter ${\boldsymbol{\theta}}$. Initialize $\z_c \sim 0.001\cdot\mathcal{N}(0,1)$, for $1\leq c\leq C$
		\While {not at end of training}
		\ForEach {mini-batch $\left\{ \left( \mathbf{x}_j, y_j\right) \right\}_{j=1}^{m}$}
            \State Initialize $l^c_{Inv}\leftarrow 0$, for $1\leq c\leq C$
		\For {$j=1, 2,\ldots, m$}
		\State $\xa_{j} \leftarrow \Call{PGDAttack}{\x_j, y_j, f_{\boldsymbol{\theta}}}$
		\Comment{Find PGD adversarial example}
		\State $\xia_j \leftarrow \x_j+\z_{y_j}$
		\State $l^{y_j}_{Inv} \leftarrow l^{y_j}_{Inv} + \mathcal{L}_{\text{Inv}}(\xia_j,y_j)$
		\EndFor 
		\For {$c=1,\ldots,C$}
		\State $\z_c \leftarrow
		\Pi_{\|\z_c\|_\infty\leq \epsilon'}
		\left(
		    \z_c-\alpha'\cdot\operatorname{sign}
		    \left(
		        \nabla_{\z_c}l^c_{Inv}
		    \right)
		\right)
		$
		\Comment{Update class-specific inverse adversaries}
		\EndFor
            \State Obtain $\mathbf{p}_j^{(t)}$, for $1\leq j\leq m$, by Eq.~(\ref{eq:8}) according to current epoch number $t$
            \Comment{Inverse adversary momentum}
		\State $\boldsymbol{\theta} \leftarrow \boldsymbol{\theta} - \tau \cdot \nabla_{\boldsymbol{\theta}} \left\{\sum_j \mathcal{L}_{\text{CE}}\left(f_{\boldsymbol{\theta}}\left(\xa_j \right), y_j \right) + \lambda \cdot \mathcal{L}_{\text{KL}}\left(\mathbf{p}_j^{(t)} \| f_{\boldsymbol{\theta}}\left(\xa_j \right) \right) \right\}$ 
		\EndFor
		\EndWhile \\
		\Return Inverse adversarially trained model $f_{\boldsymbol{\theta}}$. 
	\end{algorithmic}
\end{algorithm*}

\subsection{Class-specific Inverse Adversaries}
We have introduced the instance-wise inverse adversarial example in the previous section, which is effective in approximating the high-confidence region in the decision surface. However, the inverse adversary generation suffers from a high computational cost due to iterative gradient computation. In general, the instance-wise inverse adversary generation can take almost the same time as the original adversary generation. To reduce the computational cost of inverse adversary generation, we further design a \textit{Class-Specific Universal} inverse adversary generation strategy inspired by \cite{moosavi2017universal, shafahi2020universal}. The universal strategy allows examples of the same class to share a universal adversarial perturbation. In other words, each class owns a universal inverse adversarial perturbation that can be effective in approaching its high-likelihood region (lower the objective loss). In this way, we can find a shared direction to reach high-likelihood regions, which can also mitigate the individual noise between different examples. The class-specific universal adversarial perturbation $\z_c$ for class $c$ can be defined as: 

\begin{equation}
	\begin{aligned}
         \mathcal{L}_{\text{Inv}}\left(\Pi_{\mathbb{B}(\mathbf{x}^c, \epsilon')} \left(\mathbf{x}^c + \z_c \right), y^c \right) &< \mathcal{L}_{\text{Inv}}\left( \mathbf{x}^c, y^c \right) \\
	   \text{ for ``most'' } \mathbf{x}^c \sim & \mathcal{D}^c .
		\label{eq:5}
	\end{aligned}
\end{equation} 
We sample natural examples $\mathbf{x}^c$ and corresponding labels $y^c$ from dataset $\mathcal{D}^c$ of category $c$. The class-specific universal inverse perturbation $\z_c$ is effective in most of the examples from the same class $c$ for reducing the loss. Note that we keep updating class-specific inverse perturbations throughout the whole training stage. For a certain batch of data, we can obtain the updated universal inverse perturbation by solving the following optimization problem:

\begin{equation}
	\begin{aligned}
		\min\limits_{\left\| \z_c \right\|_{\infty}<\epsilon} \frac{1}{N_c}\sum_{i=1}^{N_c}\mathcal{L}_{\text{Inv}}\left( \mathbf{x}^c_{i}+\z_c, y^c_{i} \right),
		\label{eq:6}
	\end{aligned}
\end{equation} 
where $N_c$ is the number of training samples belonging to class $c$ of a batch. Specifically, we can further use PGD to solve the above optimization problem to obtain class-specific inverse adversarial perturbation $\z_c$. For time efficiency, we only conduct a single-step PGD to update the universal inverse perturbation for a certain category.

\begin{table*}[!t]
	\centering
	\caption{Comparison of our methods (UIAT) using ResNet-18 trained on CIFAR-10, CIFAR-100, and SVHN with other adversarial training methods. The $\ell_{\infty}$-norm adversarial perturbations are restricted in $\epsilon = 8/255$. We report both natural accuracy (\%) and robust accuracy (\%). The best result in each column is in \textbf{bold}.}
        \vspace{-3mm}
	\resizebox{0.99\linewidth}{!}{
	\begin{tabular}{ccccccccccccc}
		\toprule
		\multirow{2}{*}{Method} &\multicolumn{4}{c}{CIFAR-10}&\multicolumn{4}{c}{CIFAR-100}&\multicolumn{4}{c}{SVHN}\\
		\cmidrule(r){2-5} \cmidrule(r){6-9} \cmidrule(r){10-13}
		& Natural & PGD & CW & AA & Natural & PGD & CW & AA & Natural & PGD & CW & AA\\
		\midrule
		SAT \cite{MadryMSTV18} & 83.80  & 51.40 & 50.17  & 47.68 & 57.39  & 28.36 & 26.29  & 23.18 & 92.46  & 50.55 & 50.40  & 46.07\\
		TRADES \cite{zhang2019theoretically} & 82.45  & 52.21 & 50.29  & 48.88 & 54.36  & 27.49 & 24.19  & 23.14 & 90.63  & 58.10 & 55.13  & \textbf{52.62}\\
		MART \cite{wang2019improving} & 82.20 & 53.94 & 50.43  & 48.04 & 54.78  & 28.79 & 26.15  & 24.58 & 89.88  & \textbf{58.48} & 52.48 & 48.44\\
		HAT \cite{RadeM22} & 84.86  & 52.04 & 50.33 & 48.85 & 58.73  & 27.92 & 24.60  & 23.34 & 92.06 & 57.35 & 54.77  & 52.06\\
		\midrule
		\textbf{UIAT} & \textbf{85.01}  & 54.63 & 51.10  & \textbf{49.11} & 59.55  & \textbf{30.81} & \textbf{28.05}  & \textbf{25.73} & \textbf{93.28}  & 58.18 & \textbf{55.49}  & 52.45\\
		\textbf{UIAT} (One-off)& 84.98  & \textbf{54.79} & \textbf{51.29}  & 49.05 & \textbf{60.01}  & 30.49 & 27.56  & 25.45 & 93.14  & 58.30 & 55.45  & 52.49\\
		\midrule
		
	\end{tabular}
	}
        \vspace{-3.8mm}
	
	\label{tab:1}
\end{table*}

\subsection{Universal Inverse Adversarial Training}
We here show how the universal inverse adversaries can be used to devise an effective adversarial training algorithm. The universal inverse adversarial example $\xia$ can be obtained by adding the class-specific inverse perturbation to the original example $\x$. The loss function of \textit{Universal Inverse Adversarial Training} (UIAT) can be formulated as below:

\begin{equation}
	\begin{aligned}
    \mathcal{L}_\text{UIAT} \!=\! \mathcal{L}_{\text{CE}}\left(f_{\boldsymbol{\theta}}\left( \mathbf{\hat{x}} \right), y \right) \!+\! \lambda \cdot \mathcal{L}_{\text{KL}}\left( \mathbf{p}^{(t)}  \| f_{\boldsymbol{\theta}}\left( \mathbf{\hat{x}} \right) \right),
		\label{eq:7}
	\end{aligned}
\end{equation}
where $t$ denotes the current training epoch number. To mitigate the oscillations of noisy predictions throughout the training process, we design a momentum mechanism on the predicted probability of inverse adversaries via incorporating predictions from previous epochs. The momentum mechanism to obtain aggregate predicted probability $\mathbf{p}^{(t)}$ can thus be described as:

\begin{equation}
	\mathbf{p}^{(t)} = 
	\begin{cases}
		f_{\boldsymbol{\theta}}\left( \xia \right), & \text{if } t < T \\
		\gamma \cdot \mathbf{p}^{(t-1)} + (1-\gamma) \cdot f_{\boldsymbol{\theta}}\left( \xia \right) , & \text{if } t \geq T
	\end{cases}
	\label{eq:8}
\end{equation}
where $\gamma$ is the momentum factor. Note that we start to enable the inverse adversary momentum at epoch $T$ to stabilize the training process. The main reason is that the learned representation is unstable during the early training period. Our UIAT method can thus bridge the gap between adversarial examples and the high-likelihood region of their belonging classes for robustness enhancement. The pseudo-code of our UIAT is provided in Algorithm \ref{alg:1}. We can easily obtain standard IAT by replacing universal inverse adversaries with instance-wise ones. (See Appendix \ref{appendix:B1})

To further reduce the computational overhead, we provide a one-off strategy, which only conducts the inverse adversary generation in a certain epoch $T'$ instead of generating inverse adversaries throughout the whole training process. Before epoch $T'$, we replace $\mathbf{p}^{(t)}$ in Equation (\ref{eq:7}) with $f_{\boldsymbol{\theta}}\left( \x \right)$ for adversarial training, which is similar to \cite{wang2019improving}. Afterward, we keep replacing $\mathbf{p}^{(t)}$ with the temporary probability $\mathbf{p}^{(T')}$ in following epochs $t > T'$. More details of our one-off strategy are given in Appendix \ref{appendix:B2}.

\section{Experiments}
\label{sec:experiment}
In this section, we conduct extensive experiments to demonstrate the effectiveness and generalizability of our method. We first introduce our experimental settings, including datasets and implementation details. Next, we compare our method with state-of-the-art adversarial training methods in various settings, demonstrating the superiority of our inverse adversarial training. Moreover, we show that our method can be combined with single-step adversarial training methods, which meaningfully increases their performance at only a small additional cost.

\subsection{Experimental Setups}
\paragraph{Datasets.}
We conduct experiments on three standard datasets: CIFAR-10, CIFAR-100 \cite{krizhevsky2009learning}, and SVHN \cite{netzer2011reading}. Details of datasets are provided in Appendix \ref{appendix:A1}.
\vspace{-3mm}
\paragraph{Implementation details.}
Following the setting on RobustBench \cite{croce2020robustbench}, we use ResNet-18 \cite{he2016deep}, Pre-activation ResNet-18 (PRN-18) \cite{he2016identity}, and Wide-ResNet-28-10 (WRN-28-10) \cite{ZagoruykoK16} as the target networks. For training without extra data, we set the number of epochs to 100 for CIFAR10/100 \cite{krizhevsky2009learning}, and 30 for SVHN \cite{netzer2011reading}. We adopt Stochastic Gradient Descent (SGD) optimizer with Nesterov momentum factor $0.9$ \cite{Nesterov1983AMF}, cyclic learning rate schedule \cite{Smith_Super} with a maximum learning rate of $0.1$, and a weight decay factor of $5 \times 10^{-4}$. We adopt PGD method \cite{MadryMSTV18} with $10$ steps for adversary generation during the training stage. The maximum $\ell_{\infty}$-norm of adversarial perturbation is $\epsilon=8/255$, while the step size $\alpha$ is set as $2/255$ for CIFAR-10/100 and $1/255$ for SVHN following common practices. We set the inverse adversary radius as $\epsilon^{'}=4/255$. The regularization hyper-parameters $\beta$ and $\gamma$ are set to $1.0$ and $0.9$ in \Cref{eq:4} and \Cref{eq:8}. More details can be found in Appendix \ref{appendix:A2}.

\subsection{Results}
\label{sec:4_2}
\vspace{-2mm}
\paragraph{Performance of UIAT.}
We compare our proposed UIAT method with state-of-the-art adversarial training schemes as shown in Table \ref{tab:1}. We report the accuracies on natural examples as well as adversarial examples obtained using three strong adversarial attacks: PGD \cite{MadryMSTV18} with $20$ steps (step size $\alpha = 2/255$), CW \cite{carlini2017towards}, and Auto Attack (AA) \cite{croce2020reliable} for a rigorous robustness evaluation. Note that AA is a reliable and powerful ensemble attack that contains three types of white-box attack as well as a strong black-box one. Not only does our method enhance robust accuracy on these three datasets, but it also achieves a better clean accuracy, hence a smaller robustness gap. For CIFAR-100, our method significantly boosts the AA robust accuracy by nearly $2\%$ whilst improving the natural accuracy. Our superior performance on CIFAR-100 also represents the generalizability of UIAT on a more complicated dataset with more classes. In addition, we demonstrate that our UIAT with one-off inverse adversary generation can also obtain a similar performance as the standard version of UIAT. In other words, the freezing of well-learned class-specific perturbations can still facilitate the distribution alignment for robustness improvement.

\begin{table}[t]
	\centering
	\caption{Time cost comparison of adversarial training methods on CIFAR-10 dataset with different network architectures. We report the average training time (min/epoch) of these methods.}
        \vspace{-3mm}
	\begin{tabular}{c|ccc}
		\hline
		Method & ResNet-18 & WRN-28 & WRN-34  \\
		\hline
		Natural Training & 0.35 & 0.93 & 1.22   \\
		TRADES \cite{zhang2019theoretically}& 2.57 & 14.13 & 16.60    \\
		HAT \cite{RadeM22}& 4.02 & 16.88 & 18.95   \\
		\textbf{IAT}& 2.83 & 15.37 & 17.82    \\
		\textbf{UIAT}& 2.20 & 11.90 & 14.77    \\
		\textbf{UIAT} (One-off)& 1.96 & 10.74 & 13.36    \\
		\hline
	\end{tabular}
	\vspace{-3mm}
	\label{tab:2}
\end{table}

\begin{table}[!t]
	\centering
	\caption{Adversarial robustness results under different attack configurations using ResNet-18 on CIFAR-10. We present natural accuracy and (Auto-Attack) robust accuracy of different attack radii.}
        \vspace{-2mm}
	\begin{tabular}{cccc}
		\hline
		$\epsilon$&Method&Natural&Robust\\
		\hline
		\multirow{4}{*}{10/255}&TRADES \cite{zhang2019theoretically}&82.28&38.55\\
		&HAT \cite{RadeM22}&81.94&40.12\\
		&\textbf{UIAT}&82.79&40.61\\
		&\textbf{UIAT} (One-off)&82.76&41.16\\
		\cmidrule(r){1-4}
		\multirow{4}{*}{12/255}&TRADES \cite{zhang2019theoretically}&79.37&31.84\\
		&HAT \cite{RadeM22}&79.43&33.28\\
		&\textbf{UIAT}&79.50&34.32\\
		&\textbf{UIAT} (One-off)&79.30&34.61\\
		\cmidrule(r){1-4}
		\multirow{4}{*}{16/255}&TRADES \cite{zhang2019theoretically}&74.89&18.70\\
		&HAT \cite{RadeM22}&74.45&19.42\\
		&\textbf{UIAT}&74.29&21.82\\
		&\textbf{UIAT} (One-off)&74.86&21.96\\
		
		\hline
		
	\end{tabular}
        \vspace{-5mm}
	
	\label{tab:3}
\end{table}

\vspace{-3mm}
\paragraph{Computational cost comparison.}
In addition to outperforming state-of-the-art adversarial training methods on natural accuracy and robustness, our UIAT method also has a faster training speed. We compare the average training time (min/epoch) of our methods against other adversarial training methods, as presented in Table \ref{tab:2}. For a fair comparison, we conduct all the training experiments on a single NVIDIA Tesla A100 GPU with the same batch size $m=128$ on the CIFAR-10 dataset using three different network architectures. It can be seen that our one-off UIAT method has an additional $51\%$ time efficiency gain with respect to the state-of-the-art adversarial training method, \textit{i.e.}, HAT \cite{RadeM22} with ResNet-18. Note that the major time gap between IAT and UIAT comes from the difference in iteration times. IAT requires an instance-wise iterative inverse adversary generation manner, whilst UIAT only performs a single gradient descent step on each example.

\vspace{-3mm}
\paragraph{Adversarial training on large $\epsilon$.}
Besides the frequently-used attack configuration, we also train ResNet-18 with our UIAT method for larger $\epsilon$. Specifically, we report the robustness results of the one-off version of the UIAT method on CIFAR-10 under different $\ell_{\infty}$-norm radii: $10/255$; $12/255$; $16/255$. As shown in Table \ref{tab:3}, we observe that our UIAT method can achieve better robustness results while preserving a comparable natural accuracy as HAT \cite{RadeM22} when facing stronger adversarial attacks.

\begin{table}[!t]
	\centering
	\caption{Comparison of adversarial training methods using different networks on CIFAR-10/CIFAR-100 with extra training data. We report natural accuracy and (Auto-Attack) robust accuracy.}
        \vspace{-3mm}
	\resizebox{0.99\linewidth}{!}{
	\begin{tabular}{ccccc}
		\hline
		Dataset&Architecture&Method&Natural&Robust\\
		\hline
		\multirow{8}{*}{\textbf{CIFAR-10}}&\multirow{4}{*}{PRN-18}&Rebuffi \textit{et al.}\cite{rebuffi2021fixing}&83.53&56.66\\
		&&HAT \cite{RadeM22}&86.86&57.09\\
		&&\textbf{UIAT}&87.34&58.46\\
		&&\textbf{UIAT} (One-off)&87.10&58.15\\
		\cmidrule(r){2-5}
		&\multirow{4}{*}{WRN-28-10}&Rebuffi \textit{et al.}\cite{rebuffi2021fixing}&85.97&60.73\\
		&&HAT \cite{RadeM22}&88.16&60.97\\
		&&\textbf{UIAT}&88.93&61.32\\
		&&\textbf{UIAT} (One-off)&88.50&61.40\\
		\cmidrule(r){1-5}
		\multirow{8}{*}{\textbf{CIFAR-100}}&\multirow{3}{*}{PRN-18}&Rebuffi \textit{et al.}\cite{rebuffi2021fixing}&56.87&28.50\\
		&&HAT \cite{RadeM22}&61.50&28.88\\
		&&\textbf{UIAT}&62.20&29.40\\
		&&\textbf{UIAT} (One-off)&61.54&28.90\\
		\cmidrule(r){2-5}
		&\multirow{3}{*}{WRN-28-10}&Rebuffi \textit{et al.}\cite{rebuffi2021fixing}&59.18&30.81\\
		&&HAT \cite{RadeM22}&62.21&31.16\\
		&&\textbf{UIAT}&63.26&31.18\\
		&&\textbf{UIAT} (One-off)&62.45&31.43\\
		
		\hline
		
	\end{tabular}
	}
        \vspace{-5mm}
	\label{tab:4}
\end{table}

\begin{table*}[!t]
	\centering
	\caption{Robustness results of single-step adversarial training methods combined with our one-off UIAT approach on CIFAR-10. We conduct single-step adversarial training with various adversarial radii for comprehensive evaluation. We present the natural accuracy, (Auto-Attack) robust accuracy, and the average time for training an epoch.}
        \vspace{-2mm}
	\begin{tabular}{c|ccccccc}
		\hline
		\multirow{2}{*}{Method}&\multicolumn{2}{c}{$\epsilon=6/255$}&\multicolumn{2}{c}{$\epsilon=8/255$}&\multicolumn{2}{c}{$\epsilon=10/255$}&\multirow{2}{*}{Time(s)}\\
		\cmidrule(lr){2-3} \cmidrule(lr){4-5} \cmidrule(lr){6-7}
		&Natural&Robust&Natural&Robust&Natural&Robust&\\
		\hline
		N-FGSM \cite{de2022make}&84.66&56.36&80.29&48.24&75.59&41.54&48.4\\
		N-FGSM\textbf{ + UIAT}&85.53&58.21&81.85&49.84&77.85&42.77&57.3\\
		\hline
		RS-FGSM \cite{WongRK20}&86.72&55.28&84.07&46.15&86.32&0.00&32.4\\
		RS-FGSM\textbf{ + UIAT}&87.60&55.85&85.18&46.31&88.29&0.00&40.7\\
		\hline
		GradAlign \cite{andriushchenko2020understanding}&83.85&55.25&80.17&46.57&76.46&39.85&96.0\\
		GradAlign\textbf{ + UIAT}&85.52&55.46&82.31&46.74&79.11&39.56&107.8\\
		
		\hline
		
	\end{tabular}
        \vspace{-5mm}
	
	\label{tab:5}
\end{table*}

\vspace{-3mm}
\paragraph{Adversarial training with additional data.}
Following the experimental settings of \cite{gowal2020uncovering, rebuffi2021fixing, RadeM22}, we also conduct several experiments to measure the generalizability of our method with extra data. Particularly, we present the robustness results using different model architectures trained on CIFAR-10 and CIFAR-100 with 1M synthetic images produced by the Denoising Diffusion Probabilistic Model (DDPM) \cite{ho2020denoising} as the additional data. We compare our UIAT method and its one-off variant version with state-of-the-art approaches in Table \ref{tab:4}. Note that we do not apply the CutMix operation \cite{yun2019cutmix} following \cite{RadeM22}. As observed, our method obtains better robust accuracy while maintaining the same or even better natural accuracy.

\begin{table}[!t]
	\centering
		\caption{Ablation study using ResNet-18 of three component modules of UIAT for adversarially robust accuracy (\%) on CIFAR-10. }
            \vspace{-3mm}
		\begin{tabular}{c|cccccc}
			\toprule
			&  UAG & FR & IAM &Natural &PGD-20&AA\\
			\midrule
			1 & & &  & 83.97  & 53.98 & 48.33    \\
			2 & \checkmark & & & 85.19  & 53.56 & 47.63     \\
			3 & \checkmark & \checkmark & & 85.11  & 54.13 & 48.47     \\
			4 & \checkmark & & \checkmark & 84.85  & 54.29 & 48.83     \\
			\midrule
			5 & \checkmark & \checkmark & \checkmark & 85.01  & 54.63 & 49.11     \\
			
			\bottomrule
			\multicolumn{7}{l}{\footnotesize{UAG: Universal Adversary Generation.}}\\
			\multicolumn{7}{l}{\footnotesize{FR: Feature-level Regularization.}}\\
			\multicolumn{7}{l}{\footnotesize{IAM: Inverse Adversary Momentum.}}
		\end{tabular}
            \vspace{-5mm}
		\label{tab:6}
	\end{table}

\subsection{Single-Step Adversarial Training}
The computational cost for multi-step adversarial training is expensive, which has become prohibitive to adversarially train on larger models/datasets. In comparison, single-step methods try to approximate the most harmful adversarial examples with a single gradient ascent step \cite{de2022make, WongRK20, andriushchenko2020understanding, kim2021understanding} during training. Nevertheless, there still exists a considerable robustness gap between single-step adversarial training methods and multi-step ones.

In this section, we combine the one-off version of our UIAT method with state-of-the-art single-step adversarial training approaches to demonstrate the generalizability and the low time cost of our methods. For time efficiency, we set $\beta=0$ for Equation (\ref{eq:4}), which means that we only use cross-entropy loss for inverse adversary generation. More details about how to combine our UIAT method with single-step adversarial training can be found in Appendix \ref{appendix:B3}. As shown in Table \ref{tab:5}, we can observe that UIAT can serve as a plug-and-play component for boosting both natural and robust accuracy. Moreover, we show that our method can effectively adapt to various adversarial training radii for better performance. The additional computational cost for the UIAT method is also acceptable. For instance, in the case of N-FGSM \cite{de2022make}, our method can further improve nearly $1.5\%$ for both natural accuracy and adversarially robust accuracy ($\epsilon=8/255$) with only about an additional $9$ seconds time cost for each training epoch.

\section{Analysis}
\subsection{Ablation Study}
In this section, we thoroughly investigate the contributions of three components in our UIAT method: 1) Universal Adversary Generation (UAG) in Equation (\ref{eq:6}), 2) Feature-level Regularization (FR) in Equation (\ref{eq:4}), and 3) Inverse Adversary Momentum (IAM) in Equation (\ref{eq:8}). We report both natural accuracy and robust accuracy on CIFAR-10 using ResNet-18 during the ablation study in \Cref{tab:6}. 

Our baseline method (The first row in \Cref{tab:6}) is the instance-wise Inverse Adversarial Training (IAT), which has already achieved a competitive robustness performance compared to other methods. It can be seen that the universal inverse adversary generation can effectively improve the performance on natural accuracy, while the robust accuracy slightly drops. Both the feature-level regularization and the inverse adversarial momentum contribute to enhancing the adversarially robust accuracy. We can obtain our UIAT method by integrating these three components, which can effectively improve natural accuracy and robustness.

\begin{figure}[t]
	\centering
	\begin{subfigure}[t]{0.49\linewidth} 
		\centering
		\includegraphics[width=\linewidth]{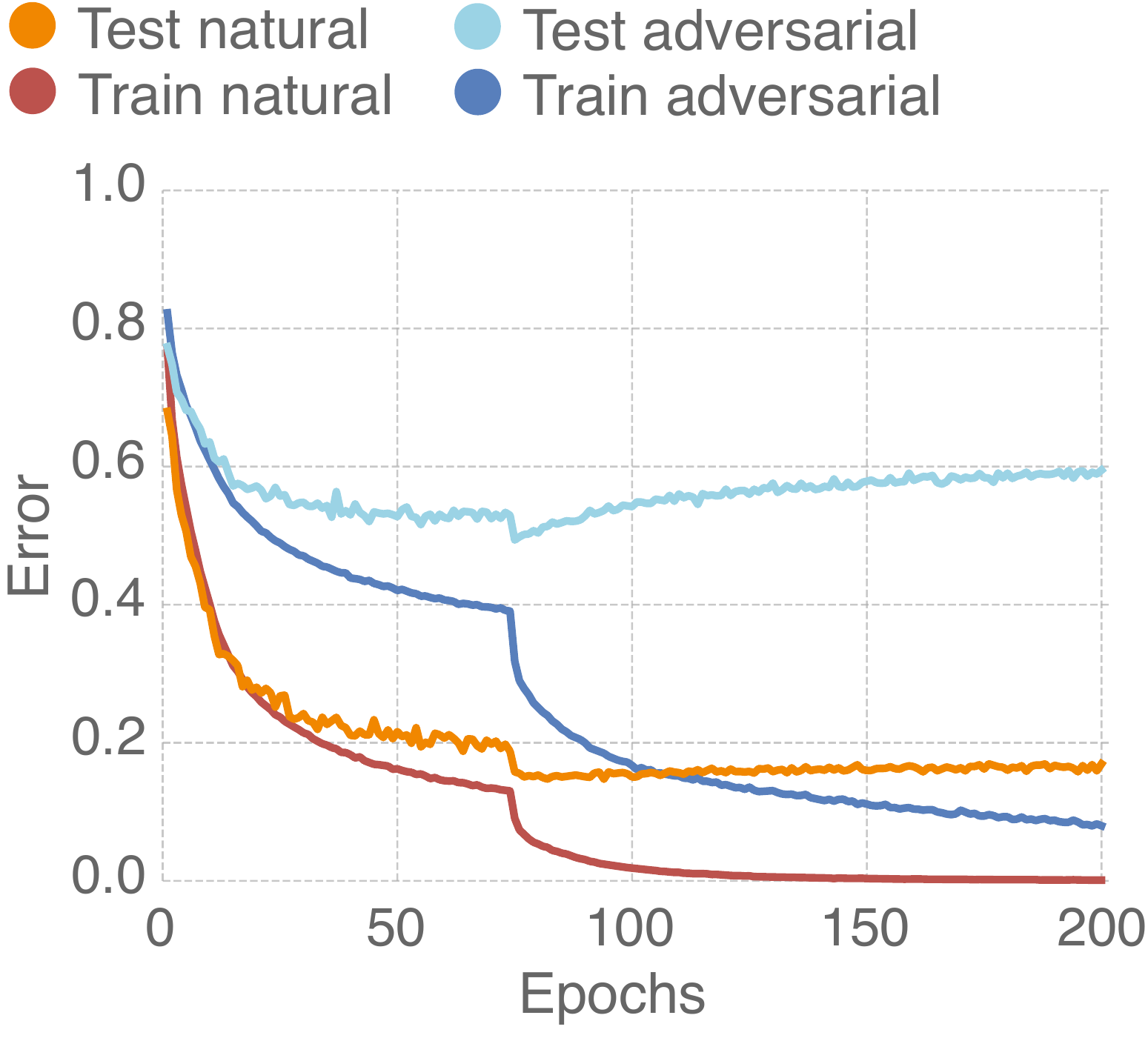}
		\caption{PGD-AT}
	\end{subfigure} 
	\begin{subfigure}[t]{0.49\linewidth}
		\centering
		\includegraphics[width=\linewidth]{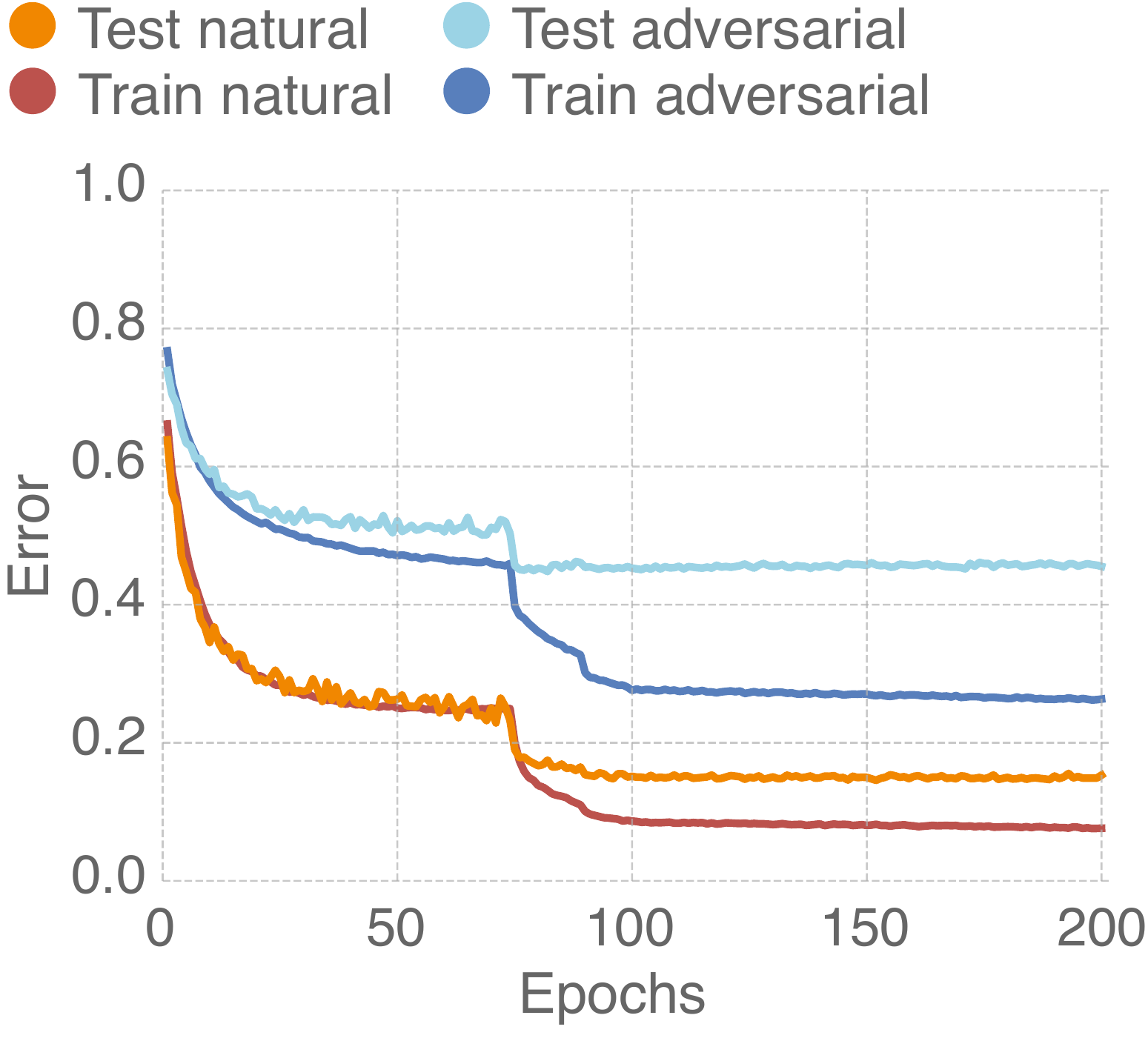}
		\caption{\textbf{UIAT} (one-off)}
	\end{subfigure} 
	\vspace{-3mm}
	\caption{The learning curves show the natural and robust accuracy (under PGD-20) on the training/test set of CIFAR-10. Note that (a) represents PGD-AT \cite{MadryMSTV18}, while (b) is our one-off UIAT method. 
	}
        \vspace{-5mm}
	\label{fig:2}
\end{figure}

\subsection{Robust Overfitting}
Recent research has demonstrated that adversarial training methods primarily suffer from the robust overfitting issue \cite{rice2020overfitting}, resulting in the robustness plunge. The robust overfitting induces an irreversible robustness drop (on the test set) after adversarial training for several epochs, especially after the learning rate decay operation. We illustrate the learning curves of standard adversarial training and our one-off version of UIAT in~\cref{fig:2}.

For better visualization, we increase the number of training epochs to $200$. We can easily observe that the PGD-based Adversarial Training (PGD-AT)~\cite{MadryMSTV18} severely suffers from the robust overfitting issue. In comparison, our one-off UIAT method can largely mitigate the robust overfitting issue, that is our method does not suffer from a sharp robustness reduction during adversarial training. It can potentially be explained by the observation made in \cite{DongXYPDSZ22}, which demonstrates that the robust overfitting comes from the large-loss data during adversarial training. However, our UIAT method implicitly regularizes the large-loss data, \textit{a.k.a.}, misclassified examples to obtain the high-likelihood region of their true classes, which can thus mitigate the robust overfitting problem. Furthermore, our inverse adversary momentum can also stabilize the training stage by mitigating the oscillations of noisy predictions.

\begin{figure}[t]
	\centering
	\includegraphics[width=0.6\linewidth]{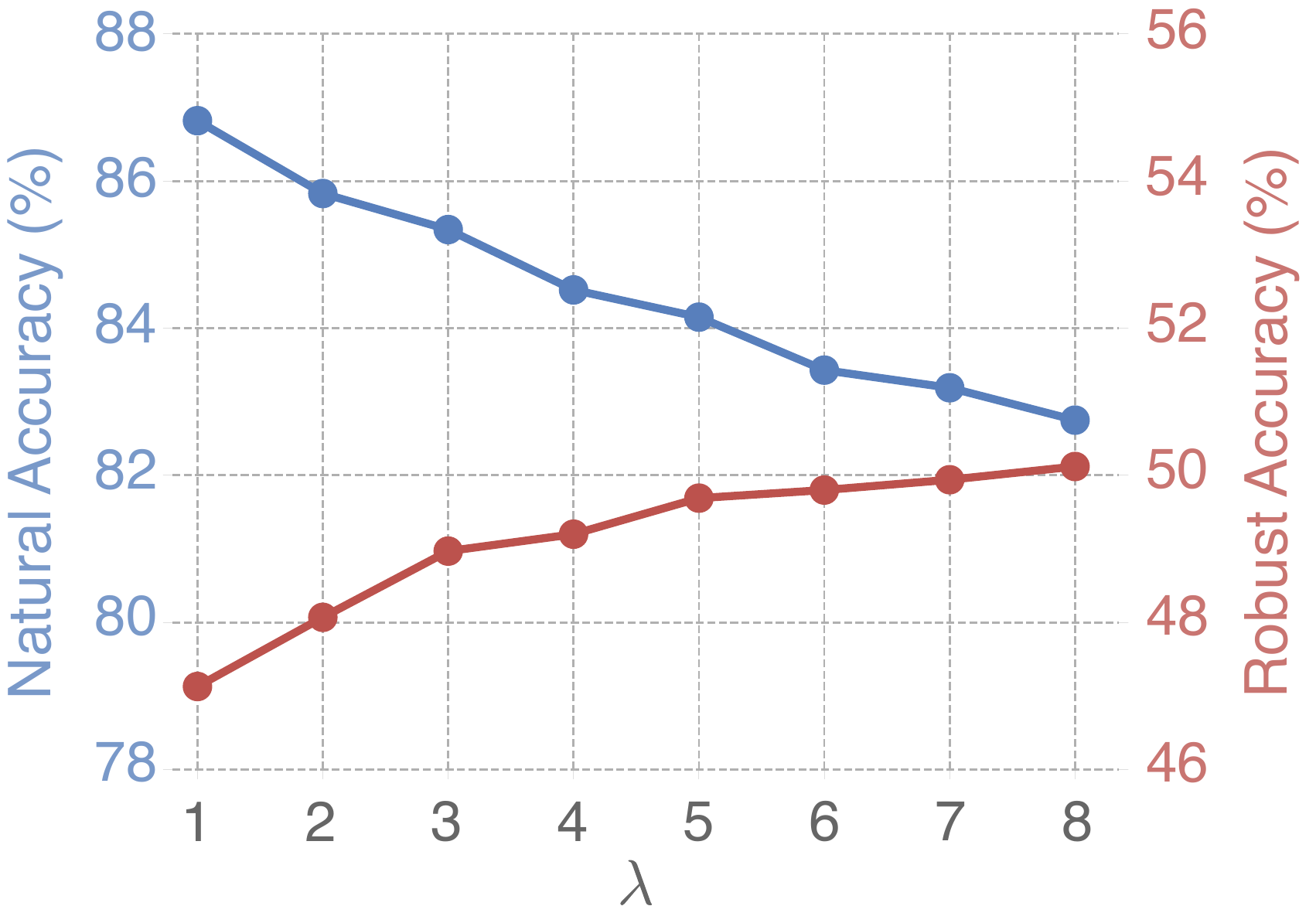}
	\vspace{-3mm}
	\caption{Parameter sensitivity of our one-off UIAT method by tuning the hyper-parameter $\lambda$. We report both natural accuracy and (Auto-Attack) robust accuracy.}
        \vspace{-2mm}
	\label{fig:3}
\end{figure}

\subsection{Trade-off}
\vspace{-1mm}
The trade-off between natural accuracy and robust accuracy during adversarial training has been widely explored \cite{zhang2019theoretically, RadeM22}. We study the effect of hyper-parameter $\lambda$, which can further induce a trade-off between robustness and natural accuracy, as shown in~\Cref{fig:3}. We can observe that the robust accuracy improves when $\lambda$ increases, while the natural accuracy decreases. Oppositely, the natural accuracy improves when we lower the $\lambda$ value. Note that our trade-off is different from \cite{zhang2019theoretically} that balances the importance of cross-entropy of natural examples and KL divergence, whilst our UIAT method optimizes the cross-entropy of adversarial examples and the regularized distribution matching. We also provide an analysis of the effect of other hyper-parameters in Appendix \ref{appendix:D}.

\begin{figure}[t]
	\centering
	\begin{subfigure}[t]{0.49\linewidth} 
		\centering
		\includegraphics[width=1\linewidth]{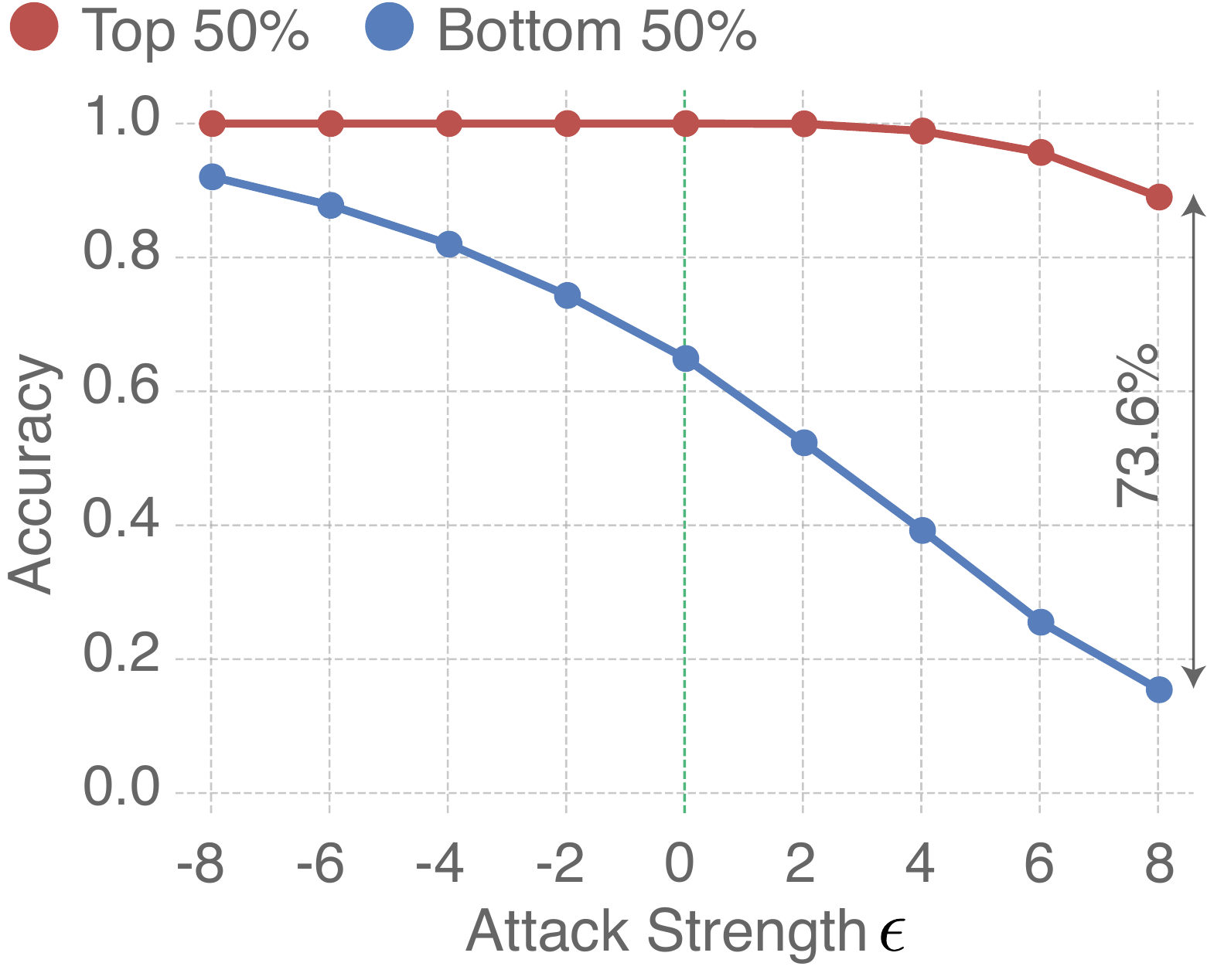}
		\caption{TRADES \cite{zhang2019theoretically}}
		\label{fig:4_1}
	\end{subfigure} 
	\begin{subfigure}[t]{0.49\linewidth}
		\centering
		\includegraphics[width=1\linewidth]{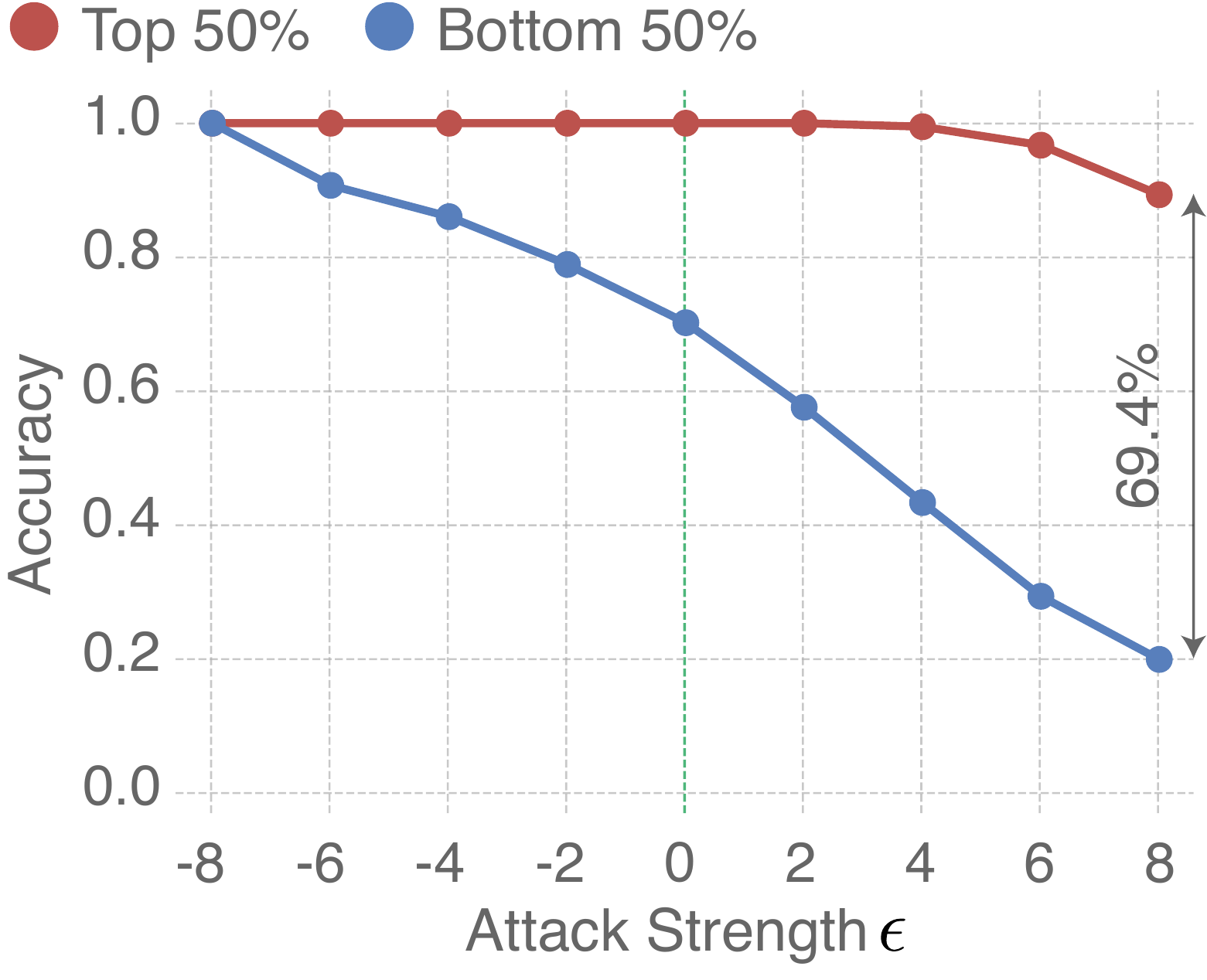}
		\caption{\textbf{UIAT} (One-off)}
		\label{fig:4_2}
	\end{subfigure} 
	\vspace{-3mm}
	\caption{Average accuracy under different attack strengths (performed by PGD-20). Note that the experimental settings are the same as \Cref{fig:1}. We further annotate the robust accuracy gap between two groups under the attack strength $\epsilon = 8/255$.}
	\label{fig:4}
	\vspace{-5mm}
\end{figure}

\begin{figure}[t]
	\centering
	\includegraphics[width=0.54\linewidth]{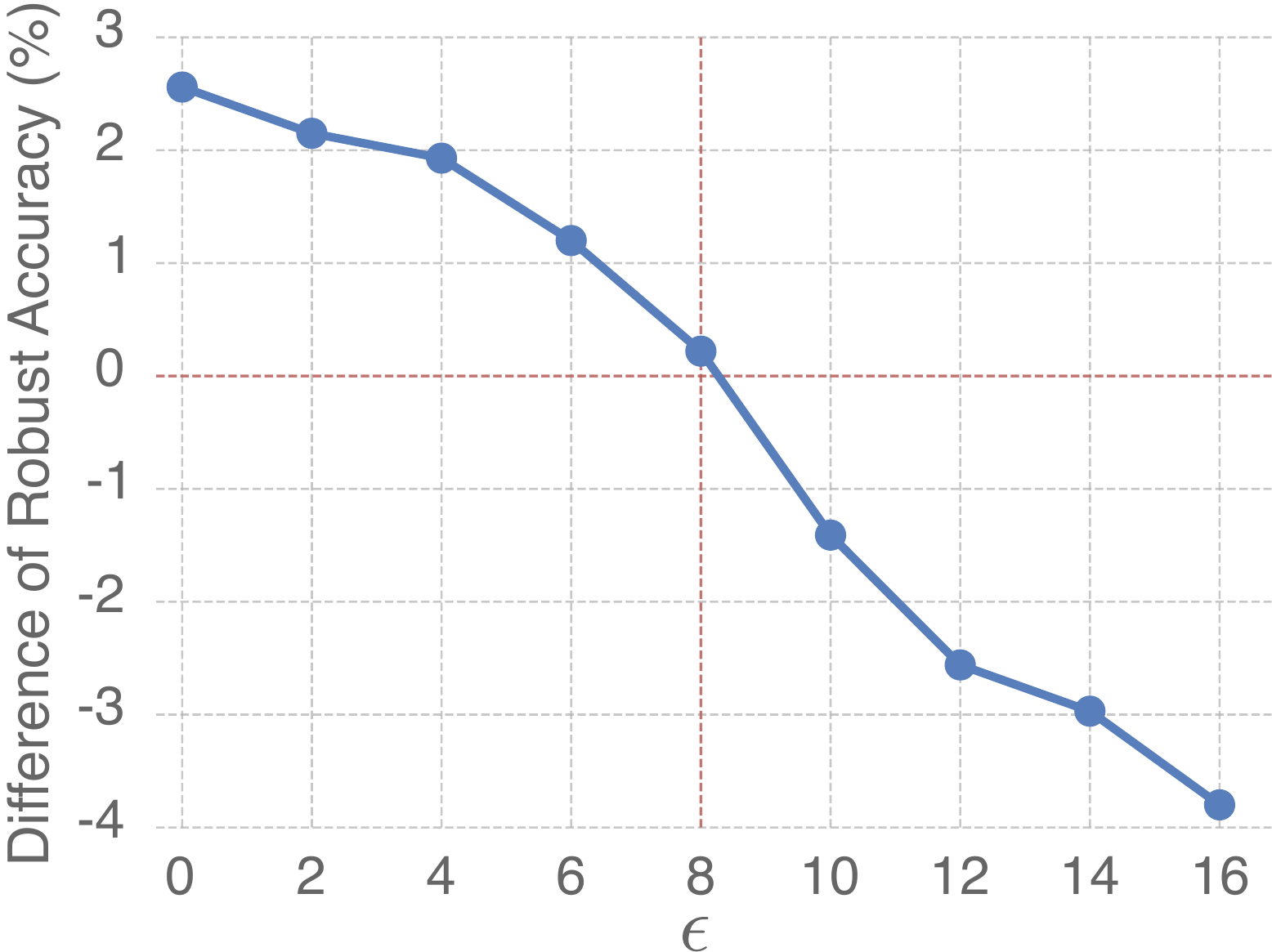}
	\vspace{-3mm}
	\caption{Difference of Auto-Attack (AA) robust accuracy under different attack strengths between our UIAT (One-off) method and TRADES \cite{zhang2019theoretically}. The \textcolor[RGB]{255,0,0}{red} lines are used for reference.}
        \vspace{-4mm}
	\label{fig:5}
\end{figure}

\subsection{Why our method is effective?}
\vspace{-1mm}
In this section, we mainly discuss the underlying reason why our method is effective. In other words, we would like to explore what we have gained from inverse adversarial training. Similar to the setting in \Cref{fig:1}, we also provide the average accuracy under different attack strengths of our UIAT method compared with TRADES \cite{zhang2019theoretically}, as shown in \Cref{fig:4}. It can be seen that our method can bridge the robust accuracy gap more effectively compared with TRADES \cite{zhang2019theoretically}. Precisely, our UIAT can effectively enhance the robust accuracy of the bottom $50\%$ group. In addition, we observe that the inverse adversarial examples of UIAT are prone to be classified correctly, which means that our robust model is easily affected by inverse adversaries. On the contrary, our robust model is less susceptible to adversarial examples compared with TRADES \cite{zhang2019theoretically}.

Furthermore, we present the comparison of Auto-Attack (AA) robust accuracy under different attack strengths between our UIAT (One-off) method and TRADES \cite{zhang2019theoretically}, as shown in \Cref{fig:5}. It can be easily observed that our method outperforms TRADES \cite{zhang2019theoretically} at weak attack strengths ($\epsilon \leq 8/255$). However, TRADES \cite{zhang2019theoretically} obtains better robustness than our method when confronted with strong adversarial perturbations ($\epsilon > 8/255$). In other words, our method sacrifices the adversarial robustness against larger visual perturbations to enhance the robustness against smaller ones. This is also in line with the definition and intuition that adversarial perturbations are visually undetectable. Recall that we can also obtain better robustness against larger perturbations when training with larger $\epsilon$ as discussed in Section \ref{sec:4_2}.

\vspace{-1mm}
\section{Conclusion}
\vspace{-1mm}
\label{sec:conclusion}
In this paper, we explore the unnecessary alignment between misclassified examples and propose a new adversarial training paradigm incorporating the inverse adversarial examples. Furthermore, we design a universal inverse adversary generation strategy to mitigate the class-wise imbalance hiding in the decision surface and accelerate our methods. Extensive experiments demonstrate that our method can efficiently obtain better robustness results without compromising natural accuracy in diverse settings on larger datasets. Moreover, we can obtain a trade-off between natural accuracy and robust accuracy to adapt to different scenarios. Our UIAT method can also be combined with state-of-the-art single-step adversarial training methods for robustness enhancement at a low cost. Finally, we analyze the reason why our method is effective and verify that our UIAT method can potentially bridge the accuracy gap between high-accuracy examples and low-accuracy examples, thus benefiting the robustness.

{\small
\bibliographystyle{ieee_fullname}
\bibliography{egbib}

\begin{thebibliography}{10}\itemsep=-1pt

\bibitem{andriushchenko2020understanding}
Maksym Andriushchenko and Nicolas Flammarion.
\newblock Understanding and improving fast adversarial training.
\newblock {\em Advances in Neural Information Processing Systems},
  33:16048--16059, 2020.

\bibitem{BaiL0WW21}
Tao Bai, Jinqi Luo, Jun Zhao, Bihan Wen, and Qian Wang.
\newblock Recent advances in adversarial training for adversarial robustness.
\newblock In {\em Proceedings of the Thirtieth International Joint Conference
  on Artificial Intelligence, {IJCAI}}, pages 4312--4321, 2021.

\bibitem{carlini2017adversarial}
Nicholas Carlini and David Wagner.
\newblock Adversarial examples are not easily detected: Bypassing ten detection
  methods.
\newblock In {\em Proceedings of the 10th ACM workshop on artificial
  intelligence and security}, pages 3--14, 2017.

\bibitem{carlini2017towards}
Nicholas Carlini and David Wagner.
\newblock Towards evaluating the robustness of neural networks.
\newblock In {\em 2017 ieee symposium on security and privacy (sp)}, pages
  39--57. Ieee, 2017.

\bibitem{cody2017dawnbench}
Cody Coleman, Deepak Narayanan, Daniel Kang, Tian Zhao, Jian Zhang, Luigi
  Nardi, Peter Bailis, Kunle Olukotun, Chris R{\'e}, and Matei Zaharia.
\newblock Dawnbench: An end-to-end deep learning benchmark and competition.
\newblock {\em NIPS ML Systems Workshop}, 2017.

\bibitem{croce2020robustbench}
Francesco Croce, Maksym Andriushchenko, Vikash Sehwag, Edoardo Debenedetti,
  Nicolas Flammarion, Mung Chiang, Prateek Mittal, and Matthias Hein.
\newblock Robustbench: a standardized adversarial robustness benchmark.
\newblock {\em arXiv preprint arXiv:2010.09670}, 2020.

\bibitem{croce2020reliable}
Francesco Croce and Matthias Hein.
\newblock Reliable evaluation of adversarial robustness with an ensemble of
  diverse parameter-free attacks.
\newblock In {\em International conference on machine learning}, pages
  2206--2216. PMLR, 2020.

\bibitem{cui2021learnable}
Jiequan Cui, Shu Liu, Liwei Wang, and Jiaya Jia.
\newblock Learnable boundary guided adversarial training.
\newblock In {\em Proceedings of the IEEE/CVF International Conference on
  Computer Vision}, pages 15721--15730, 2021.

\bibitem{de2022make}
Pau de Jorge, Adel Bibi, Riccardo Volpi, Amartya Sanyal, Philip~HS Torr,
  Gr{\'e}gory Rogez, and Puneet~K Dokania.
\newblock Make some noise: Reliable and efficient single-step adversarial
  training.
\newblock {\em arXiv preprint arXiv:2202.01181}, 2022.

\bibitem{dong2022improving}
Junhao Dong, Yuan Wang, Jian-Huang Lai, and Xiaohua Xie.
\newblock Improving adversarially robust few-shot image classification with
  generalizable representations.
\newblock In {\em Proceedings of the IEEE/CVF Conference on Computer Vision and
  Pattern Recognition}, pages 9025--9034, 2022.

\bibitem{DongXYPDSZ22}
Yinpeng Dong, Ke Xu, Xiao Yang, Tianyu Pang, Zhijie Deng, Hang Su, and Jun Zhu.
\newblock Exploring memorization in adversarial training.
\newblock In {\em The Tenth International Conference on Learning
  Representations, {ICLR}}, 2022.

\bibitem{GoodfellowSS14}
Ian Goodfellow, Jonathon Shlens, and Christian Szegedy.
\newblock Explaining and harnessing adversarial examples.
\newblock In {\em ICLR}, 2015.

\bibitem{gowal2020uncovering}
Sven Gowal, Chongli Qin, Jonathan Uesato, Timothy Mann, and Pushmeet Kohli.
\newblock Uncovering the limits of adversarial training against norm-bounded
  adversarial examples.
\newblock {\em arXiv preprint arXiv:2010.03593}, 2020.

\bibitem{he2016deep}
Kaiming He, Xiangyu Zhang, Shaoqing Ren, and Jian Sun.
\newblock Deep residual learning for image recognition.
\newblock In {\em Proceedings of the IEEE conference on computer vision and
  pattern recognition}, pages 770--778, 2016.

\bibitem{he2016identity}
Kaiming He, Xiangyu Zhang, Shaoqing Ren, and Jian Sun.
\newblock Identity mappings in deep residual networks.
\newblock In {\em European conference on computer vision}, pages 630--645.
  Springer, 2016.

\bibitem{hendrycks2016gaussian}
Dan Hendrycks and Kevin Gimpel.
\newblock Gaussian error linear units (gelus).
\newblock {\em arXiv preprint arXiv:1606.08415}, 2016.

\bibitem{HendrycksG17a}
Dan Hendrycks and Kevin Gimpel.
\newblock Early methods for detecting adversarial images.
\newblock In {\em 5th International Conference on Learning Representations,
  {ICLR}}, 2017.

\bibitem{hendrycks2021natural}
Dan Hendrycks, Kevin Zhao, Steven Basart, Jacob Steinhardt, and Dawn Song.
\newblock Natural adversarial examples.
\newblock In {\em Proceedings of the IEEE/CVF Conference on Computer Vision and
  Pattern Recognition}, pages 15262--15271, 2021.

\bibitem{ho2020denoising}
Jonathan Ho, Ajay Jain, and Pieter Abbeel.
\newblock Denoising diffusion probabilistic models.
\newblock {\em Advances in Neural Information Processing Systems},
  33:6840--6851, 2020.

\bibitem{IzmailovPGVW18}
Pavel Izmailov, Dmitrii Podoprikhin, Timur Garipov, Dmitry~P. Vetrov, and
  Andrew~Gordon Wilson.
\newblock Averaging weights leads to wider optima and better generalization.
\newblock In {\em Proceedings of the Thirty-Fourth Conference on Uncertainty in
  Artificial Intelligence, {UAI}}, pages 876--885. {AUAI} Press, 2018.

\bibitem{jia2022adversarial}
Xiaojun Jia, Yong Zhang, Baoyuan Wu, Ke Ma, Jue Wang, and Xiaochun Cao.
\newblock Las-at: Adversarial training with learnable attack strategy.
\newblock In {\em Proceedings of the IEEE/CVF Conference on Computer Vision and
  Pattern Recognition}, pages 13398--13408, 2022.

\bibitem{kahla2022label}
Mostafa Kahla, Si Chen, Hoang~Anh Just, and Ruoxi Jia.
\newblock Label-only model inversion attacks via boundary repulsion.
\newblock In {\em Proceedings of the IEEE/CVF Conference on Computer Vision and
  Pattern Recognition}, pages 15045--15053, 2022.

\bibitem{kim2021understanding}
Hoki Kim, Woojin Lee, and Jaewook Lee.
\newblock Understanding catastrophic overfitting in single-step adversarial
  training.
\newblock In {\em Proceedings of the AAAI Conference on Artificial
  Intelligence}, pages 8119--8127, 2021.

\bibitem{krizhevsky2009learning}
Alex Krizhevsky, Geoffrey Hinton, et~al.
\newblock Learning multiple layers of features from tiny images.
\newblock 2009.

\bibitem{lee2018simple}
Kimin Lee, Kibok Lee, Honglak Lee, and Jinwoo Shin.
\newblock A simple unified framework for detecting out-of-distribution samples
  and adversarial attacks.
\newblock {\em Advances in neural information processing systems}, 31, 2018.

\bibitem{li2022collaborative}
Qizhang Li, Yiwen Guo, Wangmeng Zuo, and Hao Chen.
\newblock Collaborative adversarial training.
\newblock {\em arXiv preprint arXiv:2205.11156}, 2022.

\bibitem{liu2021probabilistic}
Feng Liu, Bo Han, Tongliang Liu, Chen Gong, Gang Niu, Mingyuan Zhou, Masashi
  Sugiyama, et~al.
\newblock Probabilistic margins for instance reweighting in adversarial
  training.
\newblock {\em Advances in Neural Information Processing Systems},
  34:23258--23269, 2021.

\bibitem{long2015fully}
Jonathan Long, Evan Shelhamer, and Trevor Darrell.
\newblock Fully convolutional networks for semantic segmentation.
\newblock In {\em Proceedings of the IEEE conference on computer vision and
  pattern recognition}, pages 3431--3440, 2015.

\bibitem{MadryMSTV18}
Aleksander Madry, Aleksandar Makelov, Ludwig Schmidt, Dimitris Tsipras, and
  Adrian Vladu.
\newblock Towards deep learning models resistant to adversarial attacks.
\newblock In {\em 6th International Conference on Learning Representations,
  {ICLR} 2018}, 2018.

\bibitem{mao2019metric}
Chengzhi Mao, Ziyuan Zhong, Junfeng Yang, Carl Vondrick, and Baishakhi Ray.
\newblock Metric learning for adversarial robustness.
\newblock {\em Advances in Neural Information Processing Systems}, 32, 2019.

\bibitem{moosavi2017universal}
Seyed-Mohsen Moosavi-Dezfooli, Alhussein Fawzi, Omar Fawzi, and Pascal
  Frossard.
\newblock Universal adversarial perturbations.
\newblock In {\em Proceedings of the IEEE conference on computer vision and
  pattern recognition}, pages 1765--1773, 2017.

\bibitem{moosavi2016deepfool}
Seyed-Mohsen Moosavi-Dezfooli, Alhussein Fawzi, and Pascal Frossard.
\newblock Deepfool: a simple and accurate method to fool deep neural networks.
\newblock In {\em Proceedings of the IEEE conference on computer vision and
  pattern recognition}, pages 2574--2582, 2016.

\bibitem{Nesterov1983AMF}
Yurii Nesterov.
\newblock A method for solving the convex programming problem with convergence
  rate $o(1/k^2)$.
\newblock {\em Proceedings of the USSR Academy of Sciences}, 269:543--547,
  1983.

\bibitem{netzer2011reading}
Yuval Netzer, Tao Wang, Adam Coates, Alessandro Bissacco, Bo Wu, and Andrew~Y
  Ng.
\newblock Reading digits in natural images with unsupervised feature learning.
\newblock 2011.

\bibitem{piloto2022intuitive}
Luis~S Piloto, Ari Weinstein, Peter Battaglia, and Matthew Botvinick.
\newblock Intuitive physics learning in a deep-learning model inspired by
  developmental psychology.
\newblock {\em Nature human behaviour}, pages 1--11, 2022.

\bibitem{RadeM22}
Rahul Rade and Seyed{-}Mohsen Moosavi{-}Dezfooli.
\newblock Reducing excessive margin to achieve a better accuracy vs. robustness
  trade-off.
\newblock In {\em The Tenth International Conference on Learning
  Representations, {ICLR}}, 2022.

\bibitem{rebuffi2021fixing}
Sylvestre-Alvise Rebuffi, Sven Gowal, Dan~A Calian, Florian Stimberg, Olivia
  Wiles, and Timothy Mann.
\newblock Fixing data augmentation to improve adversarial robustness.
\newblock {\em arXiv preprint arXiv:2103.01946}, 2021.

\bibitem{rice2020overfitting}
Leslie Rice, Eric Wong, and Zico Kolter.
\newblock Overfitting in adversarially robust deep learning.
\newblock In {\em International Conference on Machine Learning}, pages
  8093--8104. PMLR, 2020.

\bibitem{ronneberger2015u}
Olaf Ronneberger, Philipp Fischer, and Thomas Brox.
\newblock U-net: Convolutional networks for biomedical image segmentation.
\newblock In {\em International Conference on Medical image computing and
  computer-assisted intervention}, pages 234--241. Springer, 2015.

\bibitem{salman2021unadversarial}
Hadi Salman, Andrew Ilyas, Logan Engstrom, Sai Vemprala, Aleksander Madry, and
  Ashish Kapoor.
\newblock Unadversarial examples: Designing objects for robust vision.
\newblock {\em Advances in Neural Information Processing Systems},
  34:15270--15284, 2021.

\bibitem{SamangoueiKC18}
Pouya Samangouei, Maya Kabkab, and Rama Chellappa.
\newblock Defense-gan: Protecting classifiers against adversarial attacks using
  generative models.
\newblock In {\em 6th International Conference on Learning Representations,
  {ICLR}}, 2018.

\bibitem{shafahi2019adversarial}
Ali Shafahi, Mahyar Najibi, Mohammad~Amin Ghiasi, Zheng Xu, John Dickerson,
  Christoph Studer, Larry~S Davis, Gavin Taylor, and Tom Goldstein.
\newblock Adversarial training for free!
\newblock {\em Advances in Neural Information Processing Systems}, 32, 2019.

\bibitem{shafahi2020universal}
Ali Shafahi, Mahyar Najibi, Zheng Xu, John Dickerson, Larry~S Davis, and Tom
  Goldstein.
\newblock Universal adversarial training.
\newblock In {\em Proceedings of the AAAI Conference on Artificial
  Intelligence}, volume~34, pages 5636--5643, 2020.

\bibitem{SimonyanZ14a}
Karen Simonyan and Andrew Zisserman.
\newblock Very deep convolutional networks for large-scale image recognition.
\newblock In {\em 3rd International Conference on Learning Representations,
  {ICLR}}, 2015.

\bibitem{Smith_Super}
Leslie~N. Smith and Nicholay Topin.
\newblock {Super-convergence: very fast training of neural networks using large
  learning rates}.
\newblock In {\em Artificial Intelligence and Machine Learning for Multi-Domain
  Operations Applications}, page 1100612. SPIE, 2019.

\bibitem{SzegedyZSBEGF13}
Christian Szegedy, Wojciech Zaremba, Ilya Sutskever, Joan Bruna, Dumitru Erhan,
  Ian~J. Goodfellow, and Rob Fergus.
\newblock Intriguing properties of neural networks.
\newblock In {\em 2nd International Conference on Learning Representations,
  {ICLR}}, 2014.

\bibitem{tian2018detecting}
Shixin Tian, Guolei Yang, and Ying Cai.
\newblock Detecting adversarial examples through image transformation.
\newblock In {\em Proceedings of the AAAI Conference on Artificial
  Intelligence}, volume~32, 2018.

\bibitem{wang2019improving}
Yisen Wang, Difan Zou, Jinfeng Yi, James Bailey, Xingjun Ma, and Quanquan Gu.
\newblock Improving adversarial robustness requires revisiting misclassified
  examples.
\newblock In {\em International Conference on Learning Representations}, 2019.

\bibitem{wen2016discriminative}
Yandong Wen, Kaipeng Zhang, Zhifeng Li, and Yu Qiao.
\newblock A discriminative feature learning approach for deep face recognition.
\newblock In {\em European conference on computer vision}, pages 499--515.
  Springer, 2016.

\bibitem{wong2018provable}
Eric Wong and Zico Kolter.
\newblock Provable defenses against adversarial examples via the convex outer
  adversarial polytope.
\newblock In {\em International Conference on Machine Learning}, pages
  5286--5295. PMLR, 2018.

\bibitem{WongRK20}
Eric Wong, Leslie Rice, and J.~Zico Kolter.
\newblock Fast is better than free: Revisiting adversarial training.
\newblock In {\em 8th International Conference on Learning Representations,
  {ICLR}}, 2020.

\bibitem{XieWZRY18}
Cihang Xie, Jianyu Wang, Zhishuai Zhang, Zhou Ren, and Alan~L. Yuille.
\newblock Mitigating adversarial effects through randomization.
\newblock In {\em 6th International Conference on Learning Representations,
  {ICLR}}, 2018.

\bibitem{yoon2021adversarial}
Jongmin Yoon, Sung~Ju Hwang, and Juho Lee.
\newblock Adversarial purification with score-based generative models.
\newblock In {\em International Conference on Machine Learning}, pages
  12062--12072. PMLR, 2021.

\bibitem{yun2019cutmix}
Sangdoo Yun, Dongyoon Han, Seong~Joon Oh, Sanghyuk Chun, Junsuk Choe, and
  Youngjoon Yoo.
\newblock Cutmix: Regularization strategy to train strong classifiers with
  localizable features.
\newblock In {\em Proceedings of the IEEE/CVF international conference on
  computer vision}, pages 6023--6032, 2019.

\bibitem{ZagoruykoK16}
Sergey Zagoruyko and Nikos Komodakis.
\newblock Wide residual networks.
\newblock In Richard~C. Wilson, Edwin~R. Hancock, and William A.~P. Smith,
  editors, {\em Proceedings of the British Machine Vision Conference 2016,
  {BMVC}}, 2016.

\bibitem{zemskova2022deep}
Varvara~E Zemskova, Tai-Long He, Zirui Wan, and Nicolas Grisouard.
\newblock A deep-learning estimate of the decadal trends in the southern ocean
  carbon storage.
\newblock {\em Nature communications}, 13(1):1--11, 2022.

\bibitem{zhang2019theoretically}
Hongyang Zhang, Yaodong Yu, Jiantao Jiao, Eric Xing, Laurent El~Ghaoui, and
  Michael Jordan.
\newblock Theoretically principled trade-off between robustness and accuracy.
\newblock In {\em International conference on machine learning}, pages
  7472--7482. PMLR, 2019.

\bibitem{ZhangZ00SK21}
Jingfeng Zhang, Jianing Zhu, Gang Niu, Bo Han, Masashi Sugiyama, and Mohan~S.
  Kankanhalli.
\newblock Geometry-aware instance-reweighted adversarial training.
\newblock In {\em 9th International Conference on Learning Representations,
  {ICLR}}, 2021.

\bibitem{zhao2019object}
Zhong-Qiu Zhao, Peng Zheng, Shou-tao Xu, and Xindong Wu.
\newblock Object detection with deep learning: A review.
\newblock {\em IEEE transactions on neural networks and learning systems},
  30(11):3212--3232, 2019.

\end{thebibliography}
}

\appendix
\paragraph{Appendix}
\section{Experimental Settings}
\label{appendix:A}
In this section, we provide detailed settings of used databases and our method.

\subsection{Datasets}
\label{appendix:A1}
We conduct all our experiments on CIFAR-10/100 \cite{krizhevsky2009learning} and SVHN \cite{netzer2011reading}. The CIFAR-10 dataset contains 60,000 color images with the size of 32 $\times$ 32 in 10 classes. The CIFAR-100 dataset shares the same setting as CIFAR-10, except it owns 100 classes consisting of 600 images each. In CIFAR-10/CIFAR-100 dataset, 50,000 images are for training, and 10,000 images are for testing the performance. SVHN is a dataset of street view house numbers, which includes 73,257 examples for training and 26,032 examples for evaluation. For training with additional data, we also include 1M synthetic images generated by the Denoising Diffusion Probabilistic Model (DDPM) \cite{ho2020denoising} for CIFAR-10/100 following the setting of \cite{rebuffi2021fixing, RadeM22}.

\begin{algorithm*}[tb] 
	\caption{\textbf{I}nverse \textbf{A}dversarial \textbf{T}raining (\textbf{IAT})}  
	\label{supp_alg:1} 
	\begin{algorithmic}[1] 
		\Statex {\bfseries Input:}  
		DNN classifier $f_{\boldsymbol{\theta}}$; dataset $\mathcal{D} = \{(\mathbf{x}, y)\}$ with $C$ classes; batch size m; learning rate $\tau$; radius for adversaries $\epsilon$ and inverse adversaries $\epsilon'$; iteration times $n$ and step size $\alpha'$ for inverse adversary generation; weighting factors $\lambda$, $\beta$. 
		\State Randomly initialize the network parameter ${\boldsymbol{\theta}}$
		\While {not at end of training}
		\ForEach {mini-batch $(\x, y) = \left\lbrace \left( \mathbf{x}_j, y_j\right) \right\rbrace_{j=1}^{m}$}
		\For {$j=1, 2,\ldots, m$}
            \State Initialize Inverse adversarial perturbation $\z_j \sim 0.001\cdot\mathcal{N}(0,1)$
		\State $\xa_{j} \leftarrow \Call{PGDAttack}{\x_j, y_j, f_{\boldsymbol{\theta}}}$
		\Comment{Find PGD adversarial example}
		\State $\xia_j \leftarrow \x_j+\z_j$
            \For {$t=1, 2, \ldots, n$}
		\State $\xia_j = \Pi_{\mathbb{B}(\mathbf{x}, \epsilon')} \left(  \xia_j - \alpha' \cdot \operatorname{sign}\left(  \nabla_{\xia_j}\mathcal{L}_{Inv}\left( \xia_j, y \right)  \right)  \right)$
            \Comment{Update instance-wise inverse adversaries}
            \EndFor 
		\EndFor
		\State $\boldsymbol{\theta} \leftarrow \boldsymbol{\theta} - \tau \cdot \nabla_{\boldsymbol{\theta}} \left\{\sum_j \mathcal{L}_{\text{CE}}\left(f_{\boldsymbol{\theta}}\left(\xa_j \right), y_j \right) + \lambda \cdot \mathcal{L}_{\text{KL}}\left(f_{\boldsymbol{\theta}}\left(\xia_j \right) \| f_{\boldsymbol{\theta}}\left(\xa_j \right) \right) \right\}$ 
		\EndFor
		\EndWhile \\
		\Return Inverse adversarially trained model $f_{\boldsymbol{\theta}}$. 
	\end{algorithmic}
\end{algorithm*}

\subsection{Implementation Details}
\label{appendix:A2}
Following the hyper-parameters setting from \cite{cody2017dawnbench, RadeM22}, we use Stochastic Gradient Descent (SGD) optimizer with Nesterov momentum factor $0.9$ \cite{Nesterov1983AMF} cyclic learning rate schedule \cite{Smith_Super} with the batch size of $128$, the maximum learning rate of $0.1$, and a weight decay factor of $5 \times 10^{-4}$. For training without extra data, our model is trained for 100 epochs for CIFAR-10/100 and 30 epochs for SVHN.

For training with synthetic DDPM-generated data \cite{rebuffi2021fixing} on CIFAR-10/100, we train models for $400$ CIFAR-10-equivalent epochs (the same amount of training examples as standard CIFAR-10 in an epoch) with the batch size of $512$. The original-to-generated ratio (\textit{e.g.}, a ratio of $0.3$ means that we include $7$ synthetic images for every $3$ original images) is $0.3$ for CIFAR-10 and $0.4$ for CIFAR-100. We also adopt the cyclic learning rate strategy with a maximum learning rate of $0.2$. Following the training setup from \cite{RadeM22, rebuffi2021fixing}, we use SiLU activation function \cite{hendrycks2016gaussian} with Pre-activation ResNet-18 (PRN-18) \cite{he2016identity} and Wide-ResNet-28-10 (WRN-28-10) \cite{ZagoruykoK16}. We further use model weight averaging \cite{IzmailovPGVW18} with a decay factor of $0.995$. 

For computing adversarial examples during training, we adopt the iterative Projected Gradient Descent (PGD) algorithm \cite{MadryMSTV18} on the cross-entropy loss function for $10$ steps with the step size $\alpha=2/255$ for CIAFR-10/100 and $\alpha=1/255$ for SVHN. We mainly consider the $\ell_{\infty}$-norm threat model with the maximum adversarial perturbation radius $\epsilon=8/255$. We set the inverse perturbation radius as $\epsilon'=4/255$. The iteration steps for instance-wise inverse perturbation is $5$ times with the step size of $\alpha'=2/255$, whilst we conduct single-step gradient descent on universal inverse perturbation with the step size of $\alpha'=4/255$. We choose the trade-off factor $\lambda=3.5$ for CIFAR10/100 and $\lambda=3.0$ for SVHN. The regularization hyper-parameters $\beta$ is set to $1.0$. We pick the inverse momentum factor $\gamma=0.9$ for our standard inverse adversarial training method except for the one-off setting. We do not involve the inverse momentum when adopting the one-off strategy. The momentum mechanism starts at epoch $T=75$ when training for $100$ epochs and starts at epoch $T=350$ when training for $400$ epochs. The one-off epoch choice is $T'=80$ for $100$ training epochs and $T'=320$ for $400$ training epochs.

\section{Details of Inverse Adversarial Training}
\label{appendix:B}
\subsection{Instance-wise Inverse Adversarial Training}
\label{appendix:B1}
We have introduced how to generate instance-wise inverse adversaries in the main body of this paper. In this section, we give more details about combining inverse adversarial examples with adversarial training. In general, we generate inverse adversarial perturbation for each natural example via the PGD method optimized on the inverse adversarial loss. The instance-wise Inverse Adversarial Training (\textbf{IAT}) is quite similar to Universal Inverse Adversarial Training (\textbf{UIAT}) we have introduced in detail. We can easily obtain IAT by replacing universal inverse adversaries with instance-wise inverse adversaries. We provide the pseudo-code of IAT in Algorithm \ref{supp_alg:1}.

\subsection{One-off Strategy}
\label{appendix:B2}
In this section, we provide more details about the one-off strategy and how it can be combined with our method. The one-off strategy means generating inverse adversarial examples for only one certain epoch $T'$ instead of throughout the whole training stage. During the standard inverse adversarial training, we mainly optimize cross-entropy loss of adversarial examples and Kullback–Leibler (KL) divergence between inverse adversaries and adversarial examples. However, the one-off strategy mainly focuses on the substitution of the inverse adversaries throughout the adversarial training, which can reduce the computational overhead effectively. The loss function for the One-Off version of inverse adversarial training can be formulated as below:

\begin{equation}
    \mathcal{L}^{OO}_\mathbf{IAT} \!=\! \mathcal{L}_{CE}\left(f_{\boldsymbol{\theta}}\left( \mathbf{\hat{x}} \right), y \right) \!+\! \lambda \cdot \mathcal{L}_{KL}\left( \mathbf{p}_{OO}^{(t)}  \| f_{\boldsymbol{\theta}}\left( \mathbf{\hat{x}} \right) \right),
    \label{supp_eq:1}
\end{equation}
where $\xa$ is the adversarial example. $\mathbf{p}_{OO}^{(t)}$ denotes the one-off output probability that mainly depends on the current training epoch $t$, which can be obtained by:

\begin{equation}
	\mathbf{p}_{OO}^{(t)} = 
	\begin{cases}
		f_{\boldsymbol{\theta}}\left( \x \right), & \text{if } t < T' \\
            f_{\boldsymbol{\theta}}\left( \xia \right), & \text{if } t = T' \\
		  \mathbf{p}_{OO}^{(T)} , & \text{if } t > T'
	\end{cases}
	\label{supp_eq:2}
\end{equation}
where $\xia$ denotes the inverse adversarial example and $T'$ is the only epoch for generating inverse adversarial examples. Before epoch $T'$, we replace inverse adversaries with natural examples during adversarial training, which is similar to \cite{wang2019improving}. We generate inverse adversarial examples and use them for distribution alignment during epoch $T'$. After epoch $T'$, we use the output probability of inverse adversaries at epoch $T$ instead of recomputing inverse adversarial examples. The motivation is that the feature representation tends to be stable at a later stage of training, thus we can consistently obtain the high-likelihood region with inverse adversarial examples. Therefore, it is reasonable to continue to use the previously computed inverse adversaries to represent the high-likelihood region during the current training epoch.

\begin{figure*}[t]
	\centering
	\includegraphics[width=0.8\linewidth]{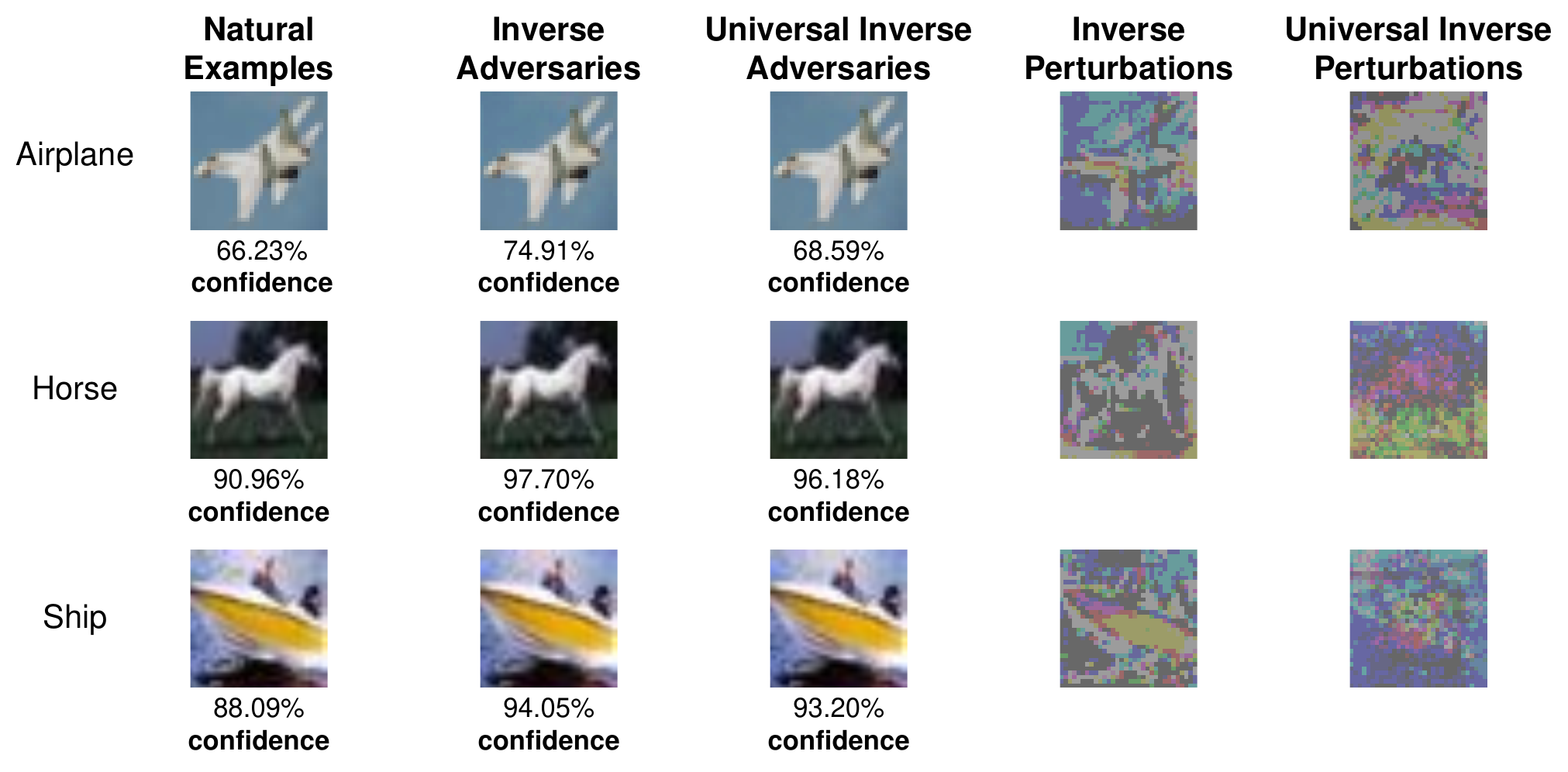}
        \vspace{-3mm}
	\caption{Visualization of both inverse adversaries and class-specific universal inverse adversaries. Their corresponding inverse adversarial perturbations are also presented.}
        \vspace{-3mm}
	\label{supp_fig:1}
\end{figure*}

\begin{figure}[t]
	\centering
        \begin{minipage}{\linewidth}
	\begin{subfigure}[t]{0.49\linewidth} 
		\centering
		\includegraphics[width=1\linewidth]{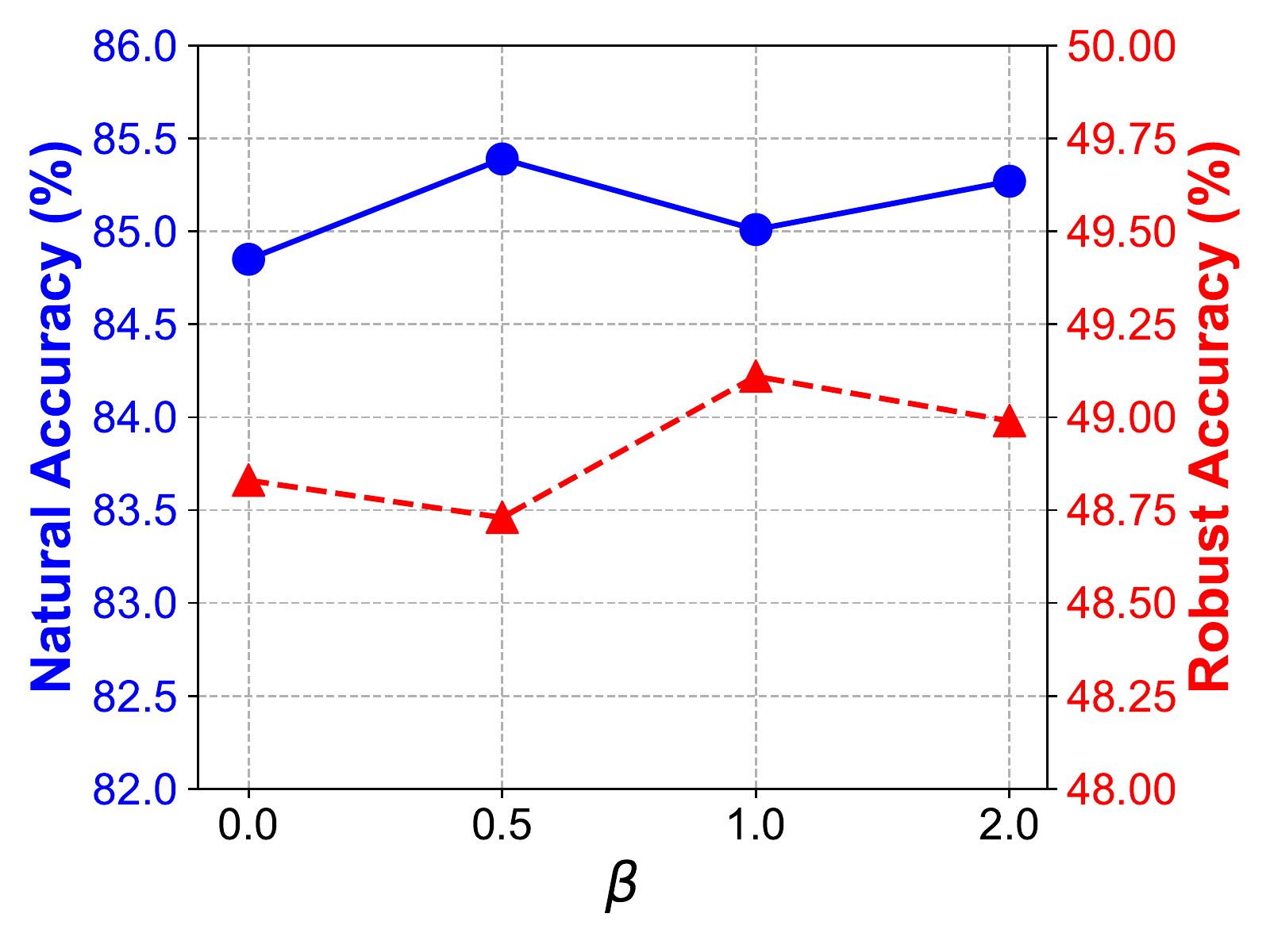}
		\caption{}
		\label{supp_fig:2_1}
	\end{subfigure} 
	\begin{subfigure}[t]{0.49\linewidth}
		\centering
		\includegraphics[width=1\linewidth]{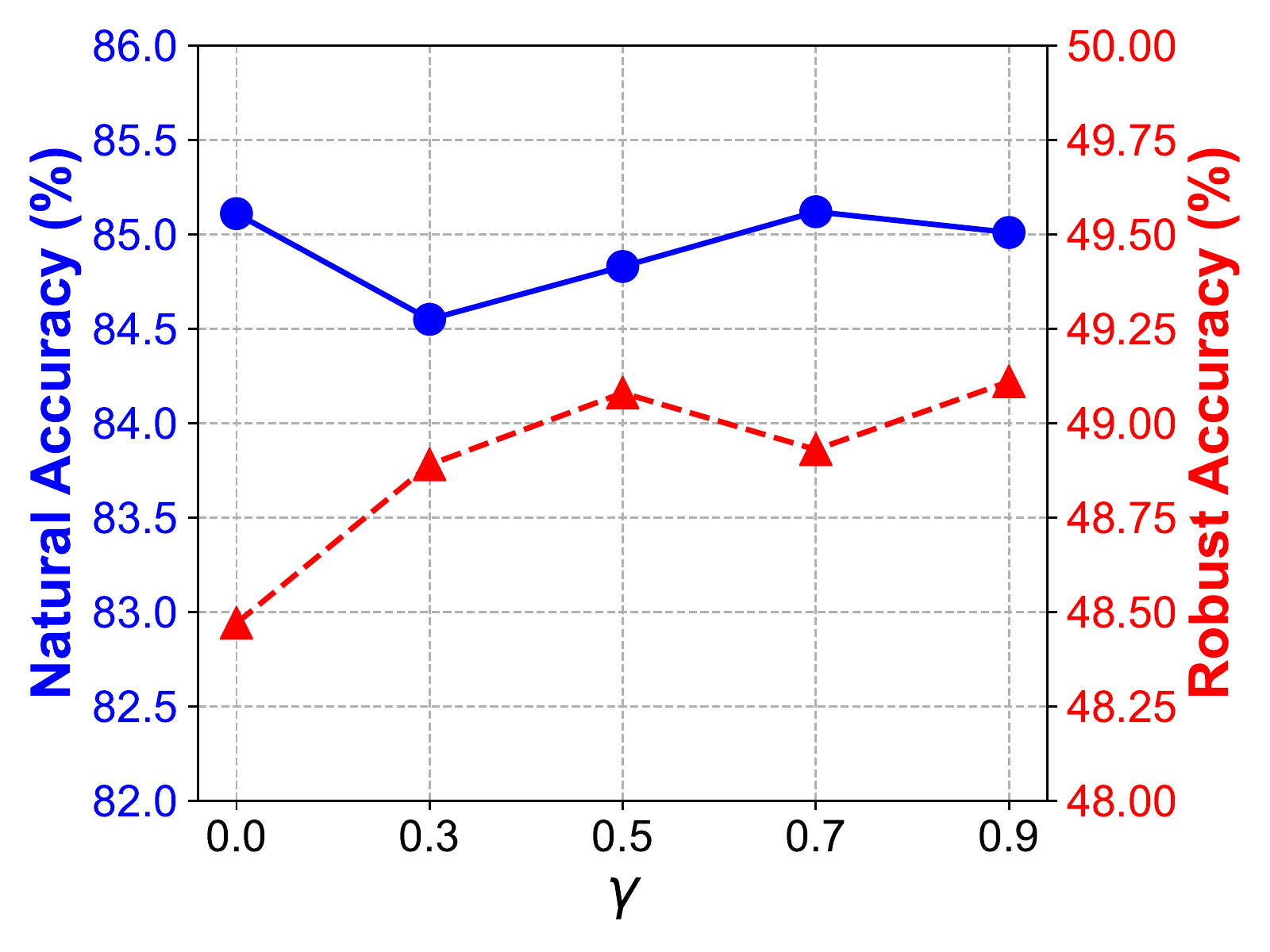}
		\caption{}
		\label{supp_fig:2_2}
	\end{subfigure} 
        \end{minipage}
        \begin{minipage}{\linewidth}
        \begin{subfigure}[t]{0.49\linewidth} 
		\centering
		\includegraphics[width=1\linewidth]{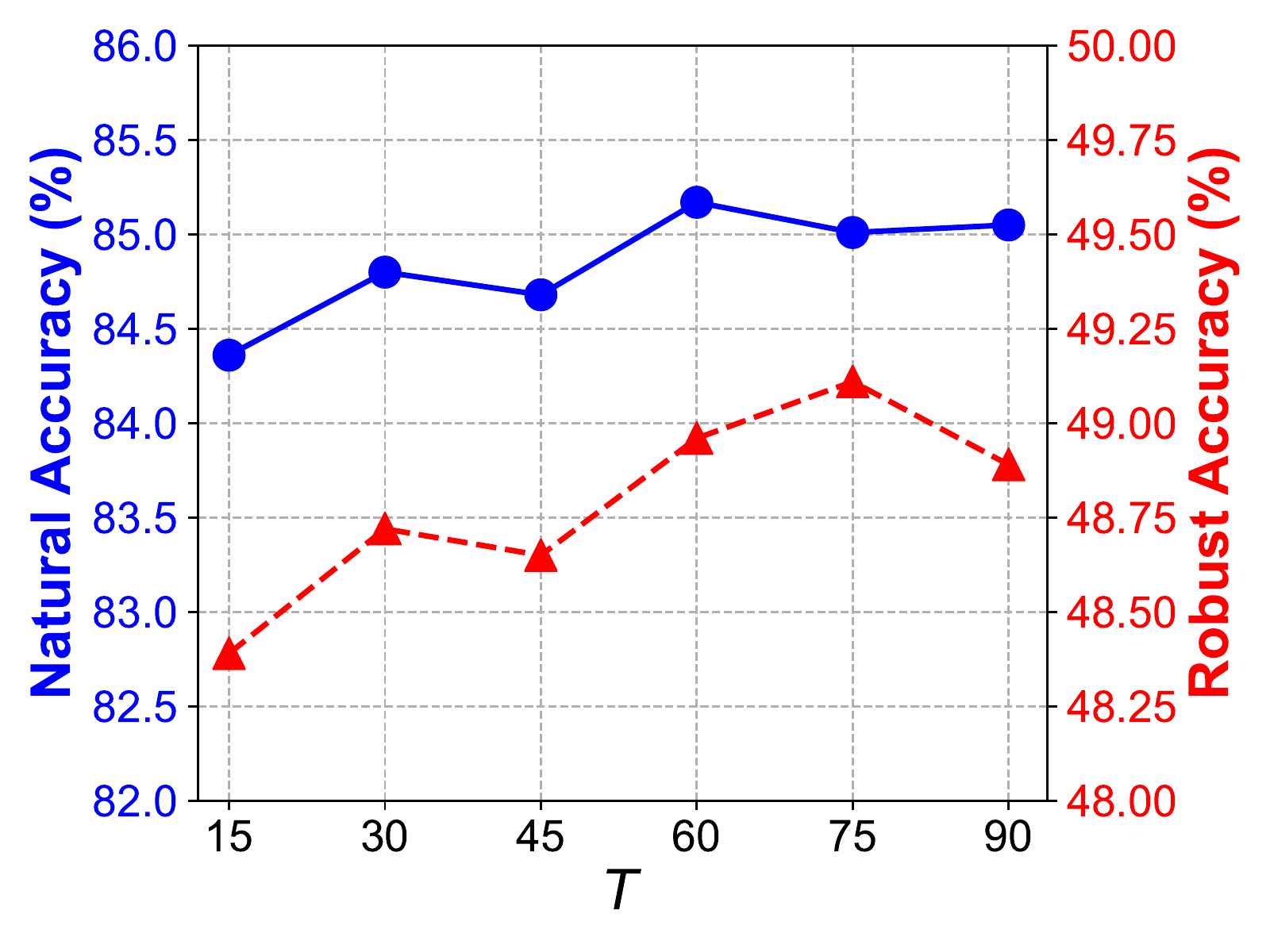}
		\caption{}
		\label{supp_fig:2_3}
	\end{subfigure} 
	\begin{subfigure}[t]{0.49\linewidth}
		\centering
		\includegraphics[width=1\linewidth]{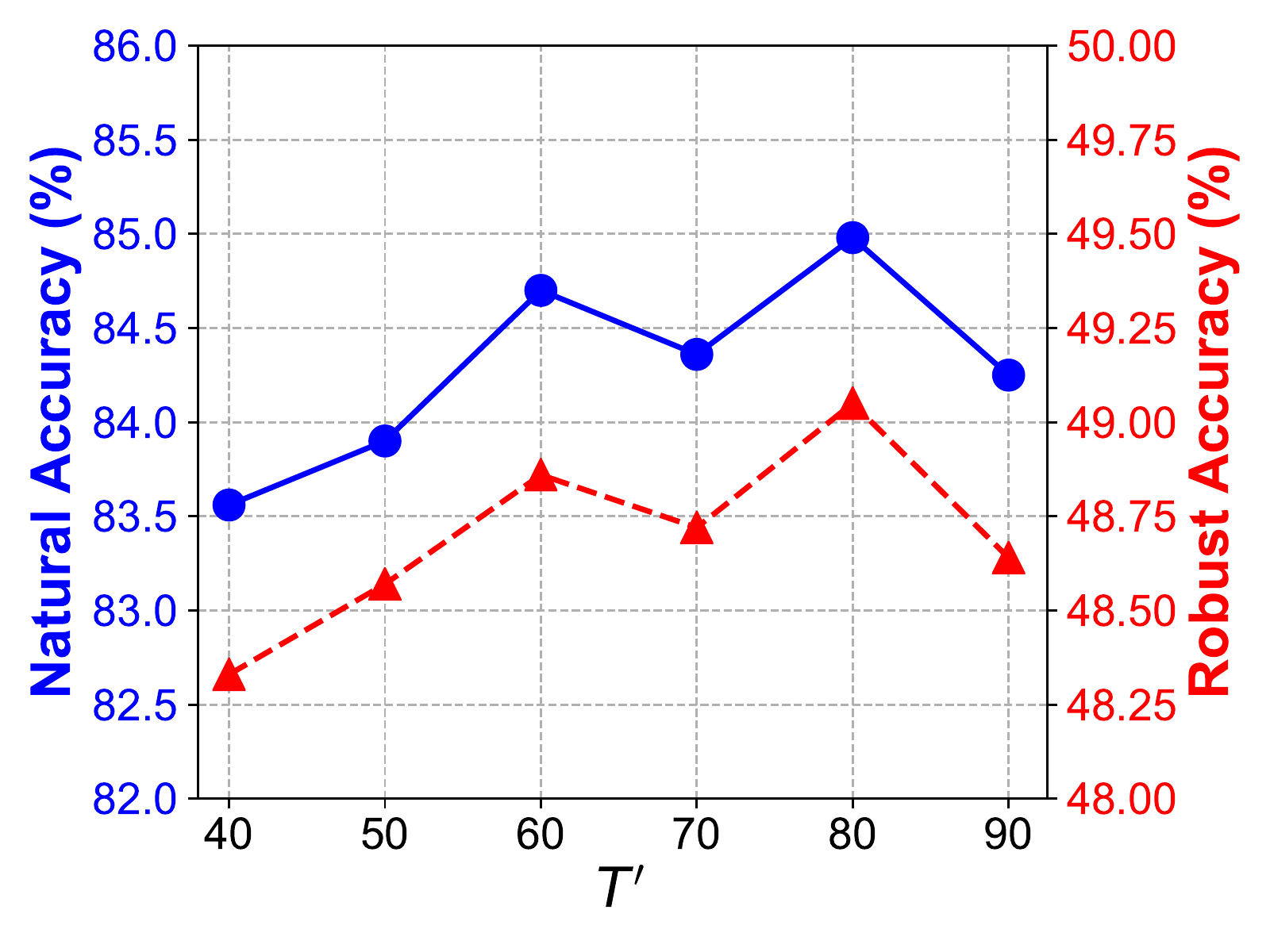}
		\caption{}
		\label{supp_fig:2_4}
	\end{subfigure} 
        \end{minipage}
	\vspace{-3mm}
	\caption{Hyper-parameter sensitivity of our UIAT method on natural accuracy and (Auto-Attack) robust accuracy using ResNet-18 on CIFAR-10. We report the hyper-parameters adjustment of $\beta$ in (a), $\gamma$ in (b). The tuning for the starting epoch of momentum $T$ is in (c), and the one-off epoch $T'$ is in (d)}
	\label{supp_fig:2}
\end{figure}

\subsection{Single-step Adversarial Training}
\label{appendix:B3}
In this section, we give more details about how our method can be combined with single-step adversarial training methods \cite{WongRK20, andriushchenko2020understanding, de2022make}. When using $\ell_{\infty}$-norm threat model, we can formalize the adversarial training \cite{MadryMSTV18} as the following min-max optimization problem:

\begin{equation}
	\begin{aligned}
		\min\limits_{\boldsymbol{\theta}} \mathbb{E}_{\left( \mathbf{x}, y\right)\sim \mathcal{D} }\left[ \max\limits_{\left\| \boldsymbol{\delta} \right\|_{\infty}<\epsilon}\mathcal{L}_{CE}\left( f_{\boldsymbol{\theta}} \left( \mathbf{x}+\boldsymbol{\delta}^{SGL}\right) , y \right)  \right],
		\label{supp_eq:3}
	\end{aligned}
\end{equation}
where $\mathcal{L}_{CE}$ is the cross-entropy loss and $\delta$ is the adversarial perturbation under the $\ell_{\infty}$-norm bound $\epsilon$. The inner maximization of adversarial training can be viewed as searching for the most harmful adversarial examples $\xa^{SGL} = \mathbf{x} + \boldsymbol{\delta}^{SGL}$. Particularly, most single-step adversarial training methods approximate the worst-case perturbation by solving the inner maximization in \Cref{supp_eq:3} with the following form:

\begin{equation}
\begin{aligned}
    \boldsymbol{\delta}^{SGL} = \psi\Big( \boldsymbol{\eta} + \alpha \cdot \operatorname{sign} \big( \nabla_{\x} \mathcal{L}_{CE}(f_\theta(\x + \boldsymbol{\eta}), y)  \big) \Big),
        \label{supp_eq:4}
\end{aligned}
\end{equation}
where $\psi$ is a projection operator onto the $\ell_{\infty}$-norm ball and $\boldsymbol{\eta}$ is drawn from a certain distribution $\boldsymbol{\Omega}$ that can be typically a uniform distribution between $[-\epsilon, \epsilon]$. When combining our UIAT method with these single-step adversarial training methods, we do not modify the inner maximization to obtain adversarial perturbations $\boldsymbol{\delta}^{SGL}$. We primarily focus on outer minimization, where we add an additional KL divergence term between universal inverse adversaries $\xia$ and adversarial examples $\xa^{SGL}$. Hence, a general form of the loss function for single-step adversarial training (outer minimization) can be defined as below:

\begin{equation}
    \mathcal{L}^{SGL}_\mathbf{IAT} = \mathcal{L}_{CE}\left(f_{\boldsymbol{\theta}}\left( \mathbf{\hat{x}} \right), y \right) + \lambda \cdot \mathcal{L}_{KL}\left( f_{\boldsymbol{\theta}}\left( \xia \right)  \| f_{\boldsymbol{\theta}}\left( \xa^{SGL} \right) \right),
    \label{supp_eq:5}
\end{equation}
where $\xia$ denotes the universal inverse adversarial example that is also obtained by single-step gradient descent on the inverse adversarial loss. Note that we do not apply the feature-level regularization during inverse adversary generation for efficiency, which means we only use cross-entropy loss for inverse adversary generation. In general, we efficiently combine our method with single-step adversarial training by paying only three additional forward propagation times and one backward propagation time.

\section{Visualization}
\label{appendix:C}
We visualize both the (universal) inverse adversarial examples and their inverse perturbations in Figure \ref{supp_fig:1}. It can be seen that class-specific universal inverse adversaries can obtain a similar inverse effect to the original inverse adversaries. These inverse examples are also visually indistinguishable from natural examples.

\section{Hyper-parameter Analysis}
\label{appendix:D}
To comprehensively analyze the contribution of each component, we report natural accuracy and robust accuracy when tuning component weights, as shown in Figure \ref{supp_fig:2}. It can be seen that enlarging the momentum factor $\gamma$ can further improve the adversarially robust accuracy. In addition, choosing the start epoch $T$ for enabling inverse adversarial momentum during the second half of training can benefit both natural accuracy and adversarial robustness. In particular, we can observe that the choice for the one-off epoch $T'$ is essential and there exists a huge performance variance when tuning this hyper-parameter. Similar to the start epoch for momentum, it is beneficial to adopt the output probability of inverse adversaries during the second half of training.

\end{document}